\documentclass[10pt,twocolumn,letterpaper]{article}

\usepackage{cvpr}
\usepackage{times}
\usepackage{epsfig}
\usepackage{graphicx}
\usepackage{amsmath}
\usepackage{amssymb}
\usepackage{graphicx}  
\usepackage{subfigure,mathrsfs,epstopdf,float,threeparttable}

\usepackage[pagebackref=true,breaklinks=true,letterpaper=true,colorlinks,bookmarks=false]{hyperref}

\cvprfinalcopy 

\def\cvprPaperID{989} 

\ifcvprfinal\fi
\begin{document}

\title{A Deep Tree-Structured Fusion Model for Single Image Deraining}

\author{Xueyang Fu$^1$,~~ Qi Qi$^1$,~~ Yue Huang$^1$,~~ Xinghao Ding$^{1*}$,~~ Feng Wu$^2$,~~ John Paisley$^3$\\
\small $^1$School of Information Science and Technology, Xiamen University, China\\
\small $^2$School of Information Science and Technology, University of Science and Technology of China, China\\
\small $^3$Department of Electrical Engineering, Columbia University, New York, NY, USA\\
{\tt\small$^*$Corresponding author: dxh@xmu.edu.cn}
}

\maketitle
\thispagestyle{empty}

\begin{abstract}
We propose a simple yet effective deep tree-structured fusion model based on feature aggregation for the deraining problem. We argue that by effectively aggregating features, a relatively simple network can still handle tough image deraining problems well. First, to capture the spatial structure of rain we use dilated convolutions as our basic network block. We then design a tree-structured fusion architecture which is deployed within each block (spatial information) and across all blocks (content information). Our method is based on the assumption that adjacent features contain redundant information. This redundancy obstructs generation of new representations and can be reduced by hierarchically fusing adjacent features. Thus, the proposed model is more compact and can effectively use spatial and content information. Experiments on synthetic and real-world datasets show that our network achieves better deraining results with fewer parameters.
\end{abstract}

\section{Introduction}
Rain can severely impair the performance of many computer vision systems, such as road surveillance, autonomous driving and consumer camera. Effectively removing rain streaks from images is an important task in the computer vision community. To address the deraining problem, many algorithms have been designed to remove rain streaks from single rainy images. Unlike video based methods \cite{garg2007vision,barnum2010analysis,bossu2011rain,santhaseelan2015utilizing,ren2017video,Jiang2017CVPR,wei2017should,Li2018VideoCVPR,Chen2018RobustCVPR}, which have useful temporal information, single image deraining is a significantly harder problem. Furthermore, success in single images can be directly extended to video, and so single image deraining has received much research attention.

In general, single image deraining methods can be categorized into two classes: model-driven and data-driven.
 Model-driven methods are designed by using handcrafted image features to describe physical characteristics of rain streaks, or exploring prior knowledge to constrain the ill-posed problem. In \cite{kim2013single}, the derained image is obtained by filtering a rainy image with a nonlocal mean smoothing filter. Several model-driven methods adopt various priors to separate rain streaks and content form rainy images. For example, in \cite{Kang2012automatic} morphological component analysis based dictionary learning is used to remove rain streaks in high frequency regions. To recognize rain streaks, a self-learning based image decomposition method is introduced in \cite{Huang2014Self}. In \cite{Luo2015Removing}, based on image patches, a discriminative sparse coding is proposed to distinguish rain streaks from non-rain content. In \cite{chen2013generalized,chang2017transformed}, low-rank assumptions are used to model and separate rain streaks. In \cite{Wang2017Hierarchical},
the authors use a hierarchical scheme combined with dictionary learning to progressively remove rain and snow. In \cite{gu2017joint}, the authors utilize convolutional analysis and synthesis sparse representation to extract rain streaks. In \cite{zhu2017joint}, three priors are explored and combined into a joint optimization process for rain removal.

\begin{figure}[t]
\begin{center}
\subfigure[Rainy image]{\includegraphics[width = 1.6in]{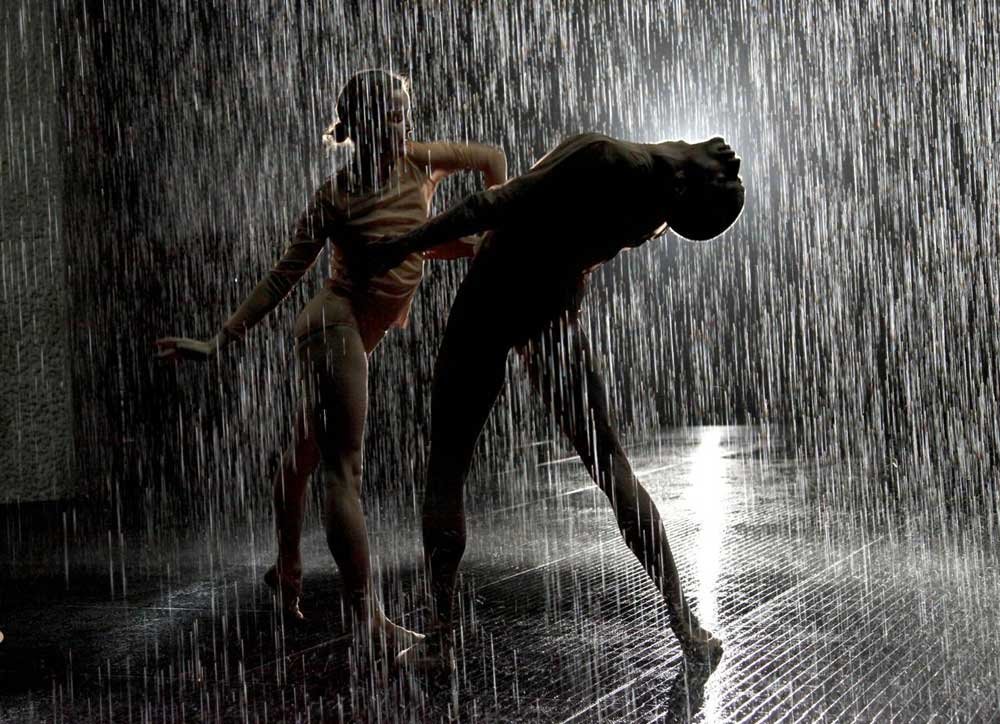}}
\subfigure[Our result]{\includegraphics[width =1.6in]{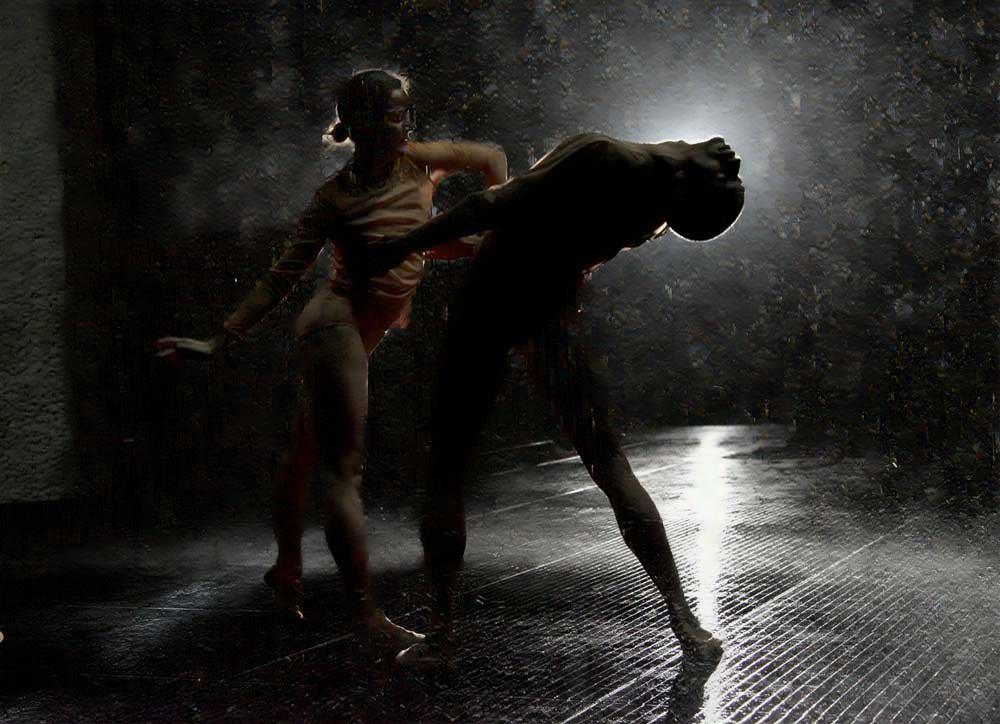}}
\caption{A deraining example of real-world image under extreme rainy conditions. Our network contains 35,427 parameters.}
\label{fig.real00}
\end{center}
\end{figure}

\begin{figure*}[th!]
\centering
{\includegraphics[width = 6.5in]{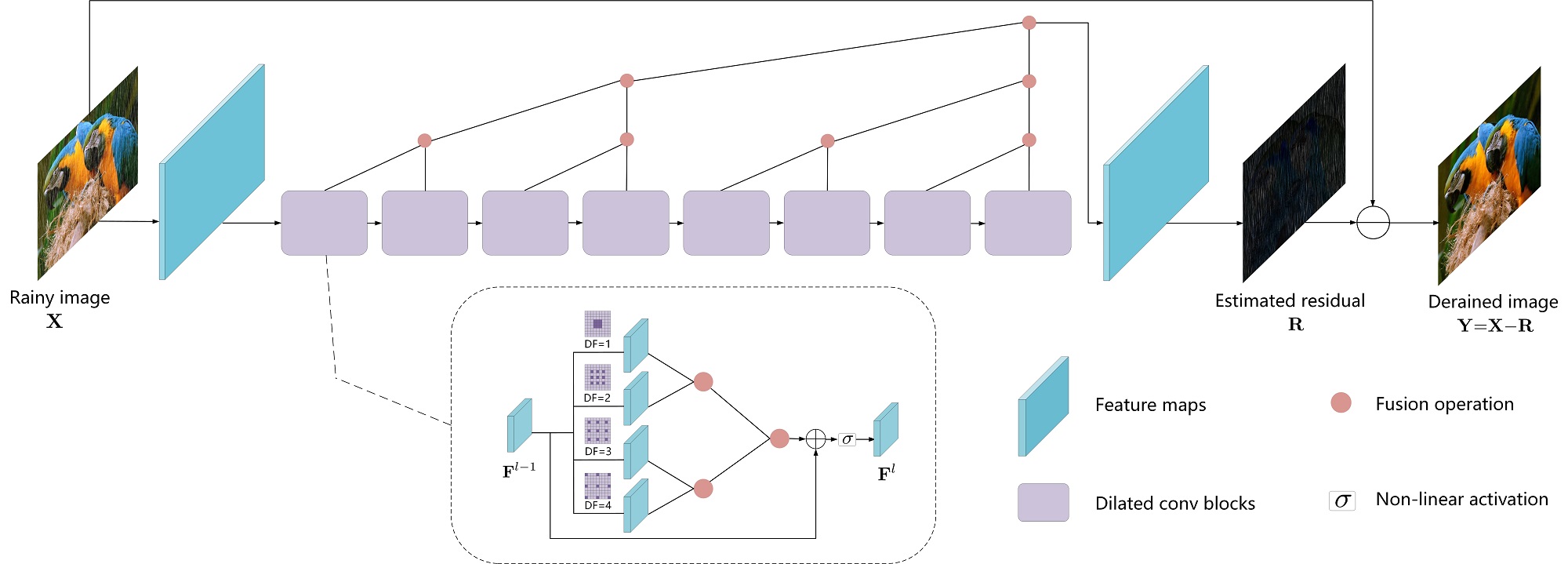}}
\caption{The framework of the proposed deep tree-structured fusion model for single image deraining. Our network contains eight dilated convolution blocks. $DF$ indicates the dilated factor. The fusion operation is expressed in equation (\ref{eq.fuse}).}\label{fig.framework}
\end{figure*}

Recently, data-driven methods using deep learning have dominated high-level vision tasks \cite{he2016deep,huang2017densely} and low-level image processing \cite{eigen2013restoring,dong2016image,tai2017memnet,Pan2018CVPR,Qian2018CVPR}. The first deep learning method for rain streaks removal was introduced by \cite{fu2017clearing}, where the authors use domain knowledge and train the network on high-frequency parts to simplify the learning processing. This method was improved in \cite{fu2017removing} by combining ResNet \cite{he2016deep} and a global skip connection. Other methods focus on designing advanced network structure to improve deraining performance. In \cite{Yang2017Deep}, a recurrent dilated network with multi-task learning is proposed for joint rain streaks detection and removal. In \cite{Li2018Recurrent}, the recurrent neural network architecture is adopted and combined with squeeze-and-excitation (SE) blocks \cite{Hu2018CVPR} for rain removal.
In \cite{zhang2018density}, a density aware multi-stream dense CNN, combined with generative adversarial network \cite{goodfellow2014generative,zhang2017image}, is proposed to jointly estimate rain density and remove rain.

Deep learning methods can focus on incorporating domain knowledge to simplify the learning process with generic networks \cite{fu2017clearing,fu2017removing} or on designing new network architectures for effective modeling representations \cite{Yang2017Deep,zhang2018density}. These works do not model the structure of the features themselves for deraining. In this paper, we show that feature fusion can improve single image deraining and reduce the number of parameters, as shown in Figure \ref{fig.real00}.

In this paper, we propose a deep tree-structured hierarchical fusion model. The proposed tree-structured fusion operation is deployed within each dilated convolutional block and across all blocks, and can explore both spatial and content information. The proposed network is easy to implement by using standard CNN techniques and has far fewer parameters than typical networks for this problem.

\section{Proposed method}
Figure \ref{fig.framework} shows the framework of our proposed hierarchical network. We adopt the multi-scale dilated convolution as the basic operation within each network block to learn multi-scale rain structures. Then, a tree-structured fusion operation within and across blocks is designed to reduce the redundancy of adjacent features. This operation enables the network to better explore and reorganize features in width and depth. The direct output of the network is the residual image, which is a common modeling technique used in existing deraining methods \cite{fu2017removing,Yang2017Deep} to ease learning. The final derained result the difference between the estimated residual and the rainy image. We describe our proposed architecture in more detail below.

\subsection{Network components}
Our proposed network contains three basic network components: one feature extraction layer, eight dilated convolution blocks and one reconstruction layer. The feature extraction layer is designed to extract basic features from the input color image. The operation of this layer is defined as by
\begin{align}\label{eq.eq1}
{{\bf{F}}^l} = \sigma ({{\bf{W}}^l}*{\bf{X}} + {{\bf{b}}^l}), ~~~l = 1,
\end{align}
where $\bf{X}$ is the input rainy image, $\bf{F}$ is the feature map, $l$ indexes layer number, $*$ indicates the convolution operation, $\bf{W}$ and $\bf{b}$ are the parameters in the convolution, and
$\sigma(\cdot)$ is the non-linear activation.

Different from typical image noise, rain streaks are spatially long. We therefore use dilated convolutions \cite{yu2015multi} in the basic network block to capture this structure. Dilated convolutions can increase the receptive field, increasing the contextual area while preserving resolution by dilating the same filter to different scales. To reduce the number of parameters, in each block we use one convolutional kernel with different dilation factors. The multi-scale features within each dilated convolution block are obtained by
\begin{align}\label{eq.eq2}
{\bf{F}}_{DF}^l = {\bf{W}}_{DF}^l*{{\bf{F}}^{l - 1}} + {{\bf{b}}^l}, ~~~l = 2,...,L - 1,
\end{align}
where $DF$  is the dilation factor, ${\bf{F}}_{DF}$  is the output feature of convolution with  $DF$, and $L$ is the total number of layers. Note that the parameters ${\bf{W}}_{DF}$ and $\bf{b}$  are shared for different dilated convolutions. The multi-scale features are fused through tree-structured aggregation to generate single-scale features ${\bf{F}}_{fused}^l$. (This hierarchical operation will be detailed at the following section.) To better propagate information, we use a skip connection to generate the output of each block by
\begin{align}\label{eq.eq3}
{{\bf{F}}^l} = \sigma ({\bf{F}}_{fused}^l + {{\bf{F}}^{l - 1}}), ~~~l = 2,...,L - 1.
\end{align}
The reconstruction layer is used to generate the color residual from previous features. The final result is obtained by
\begin{align}\label{eq.eq4}
{\bf{Y}}{\rm{ = }}{\bf{X}}{\rm{ - }}{\bf{R}} = {\bf{X}}{\rm{ - (}}{{\bf{W}}^L}*{{\bf{F}}^{L - 1}} + {{\bf{b}}^L}),
\end{align}
where  $\bf{R}$ and $\bf{Y}$ are the output residual and derained image.

\subsection{Tree-structured feature fusion}
\begin{figure}[t]
\centering
\subfigure[Parallel-structured fusion]{\includegraphics[height = 1.in]{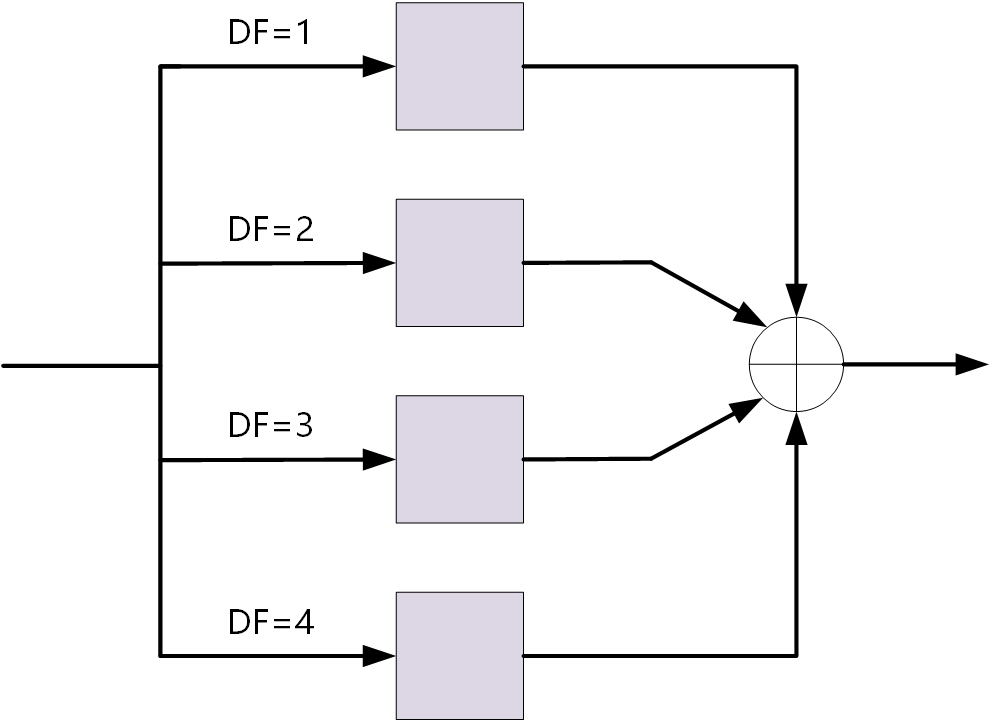}} \hspace{0.06in}
\subfigure[Tree-structured fusion]{\includegraphics[height = 1.in]{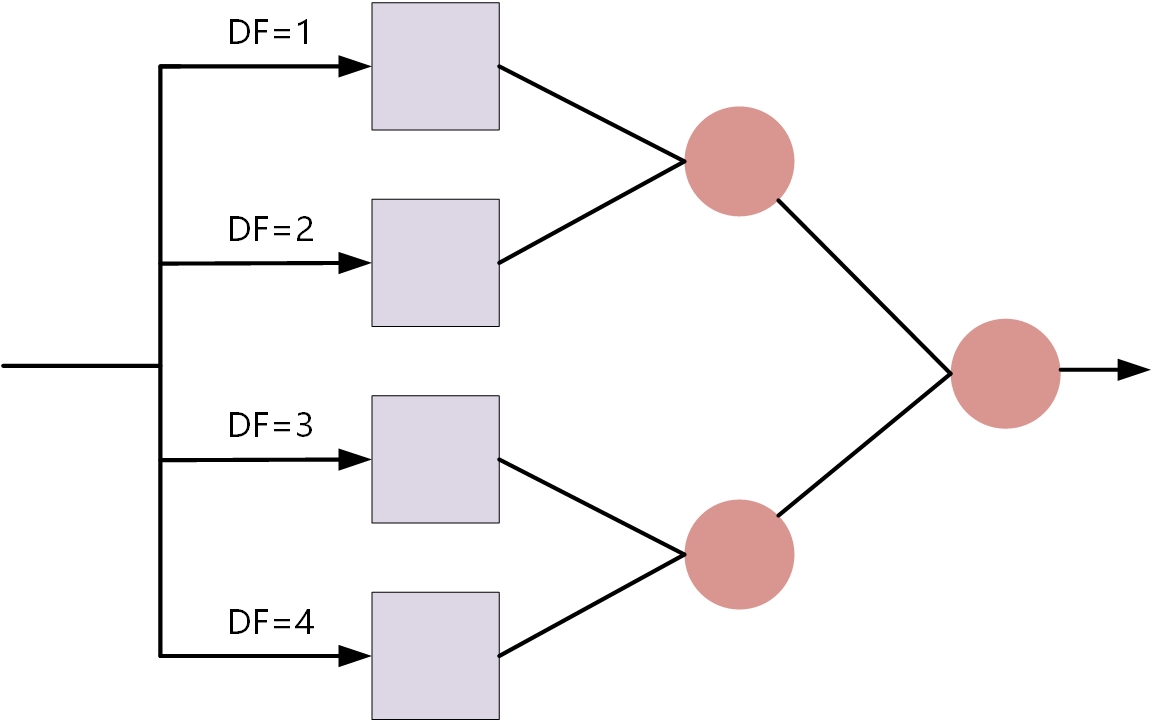}}
\caption{Comparison on fusion strategy with JORDER \cite{Yang2017Deep}. Red circles denote the operation (\ref{eq.fuse}). If all red circles are replaced by summation, (b) becomes (a).}
\label{fig.structure}
\end{figure}

In this section, we will detail our proposed tree-structured feature fusion strategy. In \cite{Yang2017Deep}, a parallel fusion structure directly added all feature maps of different dilated factors. In contrast, we design a tree-structured operation that fuses adjacent features. We use a $1 \times 1$ convolution to allow the network to automatically perform this fusion. As illustrated in Figure \ref{fig.structure}, the parallel structure of \cite{Yang2017Deep} can be seen as an instance of this tree-structured fusion in which Equation (\ref{eq.fuse}) is replaced by a summation.

We adopt and deploy this hierarchical feature fusion within each basic block and across the entire network. This allows for information propagation similar to ResNet \cite{he2016deep}, and information fusion similar to DenseNet \cite{huang2017densely}. It will also provide a sparser structure that can reduce parameters and memory usage. The fusion operation is defined as
\begin{align}
\label{eq.fuse}
{\bf{Z}} &= fuse({{\bf{Z}}_1},{{\bf{Z}}_2}) \nonumber \\
&= \sigma ({{\bf{W}}_{fuse}}*concat{\rm{(}}{{\bf{Z}}_1},{{\bf{Z}}_2}) + {\bf{b}}),
\end{align}
where ${{\bf{Z}}_1}$ and ${{\bf{Z}}_2}$  are adjacent features that have the same dimensions, $concat$ denotes the concatenation. ${{\bf{W}}_{fuse}}$ is a kernel of size $1 \times 1$ to fuse ${{\bf{Z}}_1}$  and ${{\bf{Z}}_2}$. After fusion, the generated ${\bf{Z}} $ has the same dimensions as ${{\bf{Z}}_1}$  and ${{\bf{Z}}_2}$. As shown in Figure \ref{fig.framework}, by employing this fusion operation within each block and across all blocks, the network has a tree-structured representation in both width and depth. We design this strategy based on the assumption that adjacent features contain redundant information.

\begin{figure}
\centering
\subfigure[Feature maps ($DF=3$)]{\includegraphics[height = 1in]{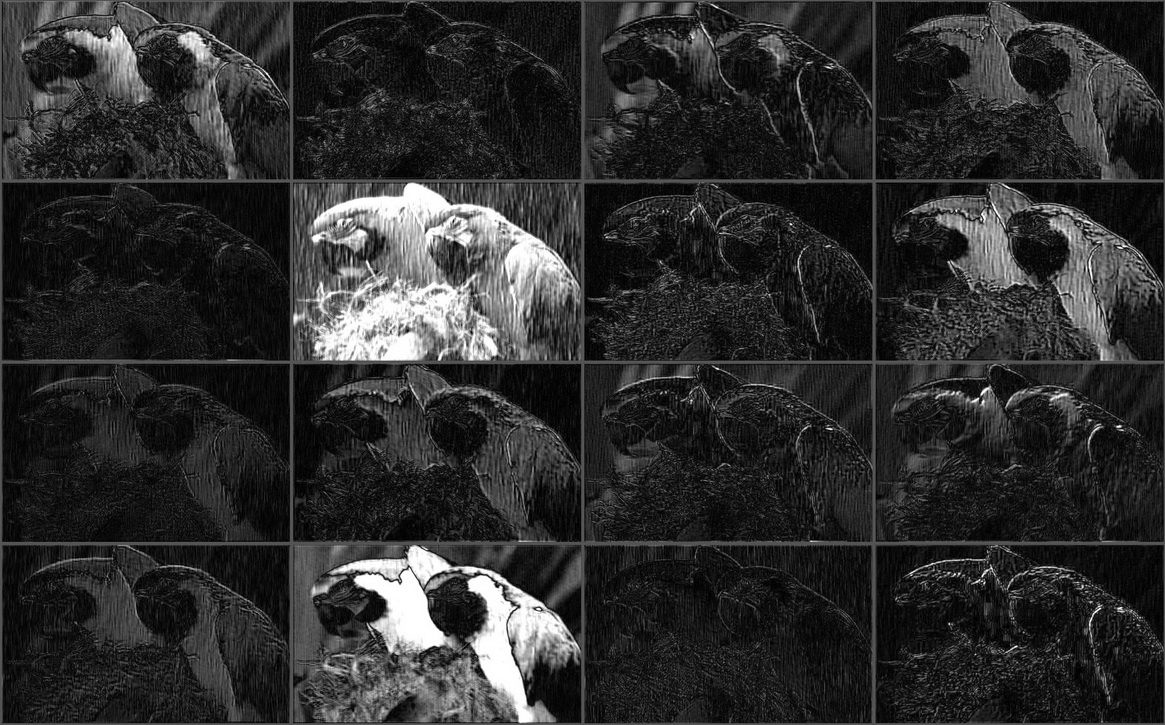}}
\subfigure[Feature maps ($DF=4$)]{\includegraphics[height = 1in]{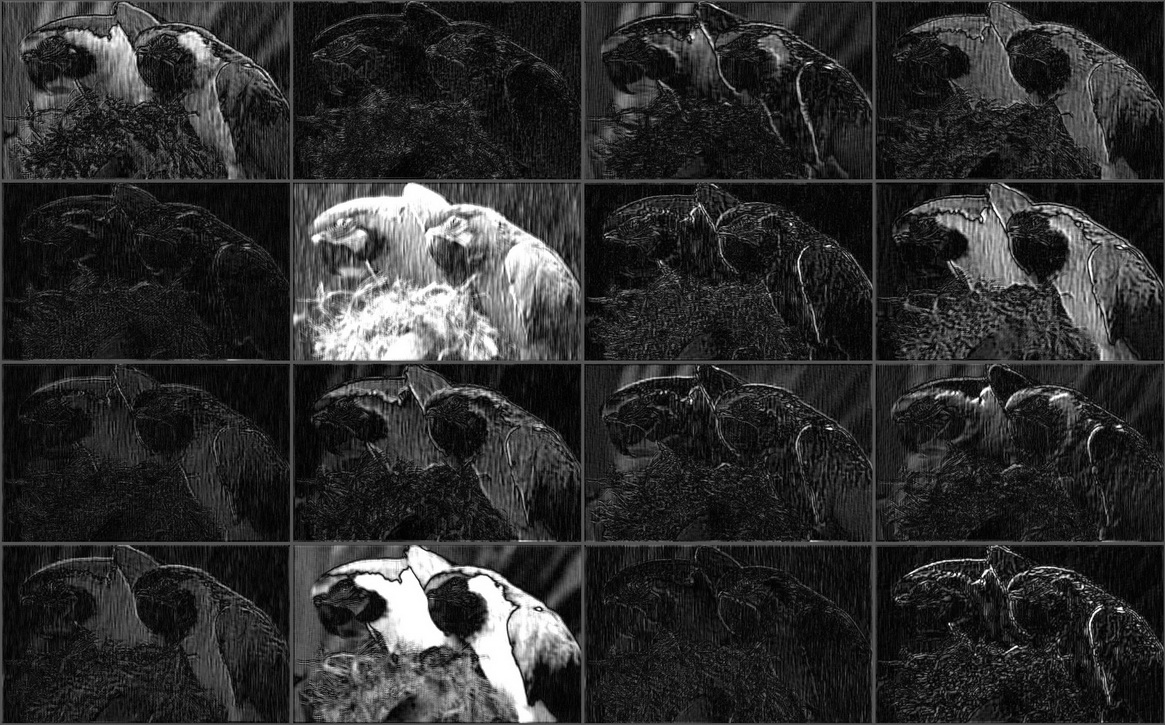}} \\
\subfigure[Directly adding (a) and (b)]{\includegraphics[height = 1in]{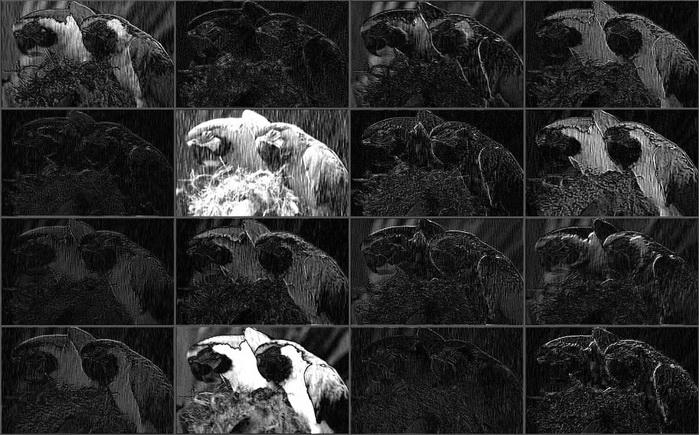}}
\subfigure[Fusing (a) and (b)]{\includegraphics[height = 1in]{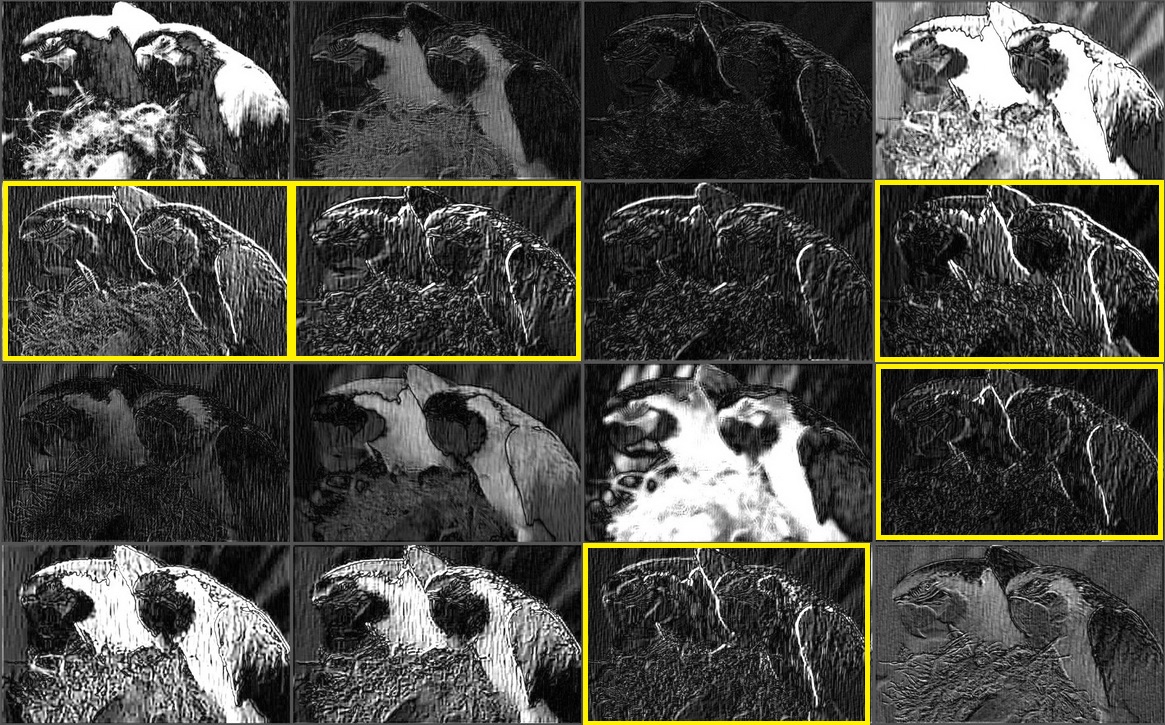}}
\caption{One example of fusion \textbf{within} the 1st block. The fused features are generated by using equation (\ref{eq.fuse}). Yellow rectangles indicates that the fused features (d) contain more effective representations for \textbf{spatial} information, which gives significant enhancement of object details and edges.}
\label{fig.dilated}
\end{figure}
To illustrate the value of this fusion operation, we show the learned, within-block feature maps in Figure \ref{fig.dilated}. In Figures \ref{fig.dilated}(a) and (b) we show the adjacent feature maps generated with dilation factors $3$ and $4$ in the first block. As can be seen, the two feature maps have similar appearance and thus contain redundant information. Figure \ref{fig.dilated}(d) shows the fused feature maps using Equation (\ref{eq.fuse}). It is clear that these fused features are more informative. Both rain and object details are highlighted.

\begin{figure}
\begin{center}
\subfigure[3rd block features]{\includegraphics[height = 1in]{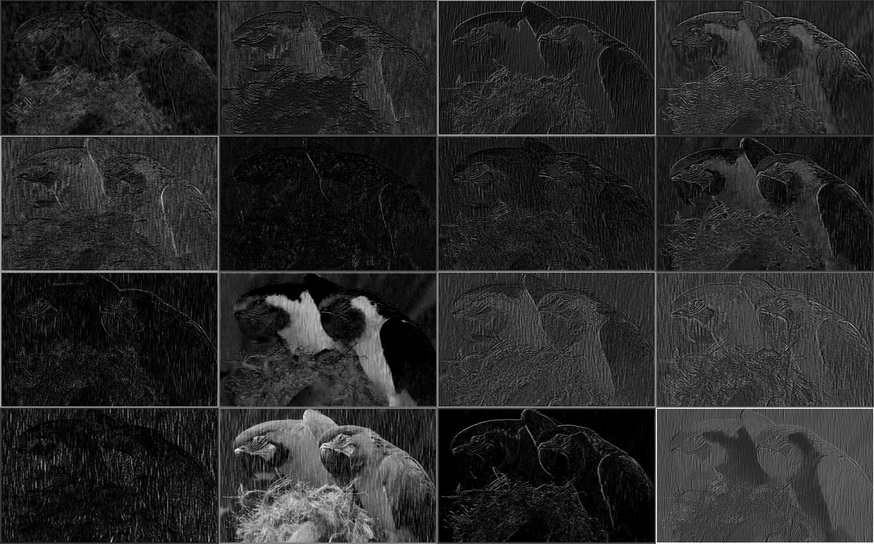}}
\subfigure[4th block features]{\includegraphics[height = 1in]{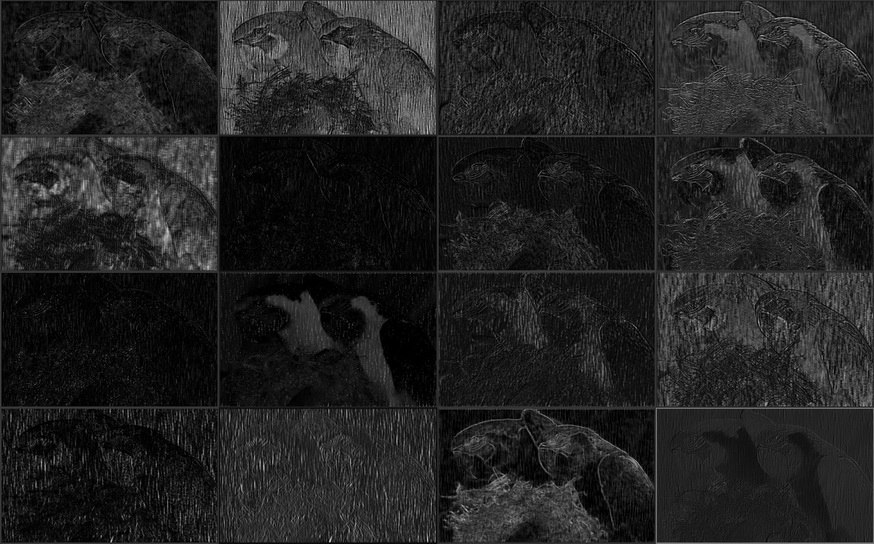}}\\
\subfigure[Directly adding (a) and (b)]{\includegraphics[height = 1in]{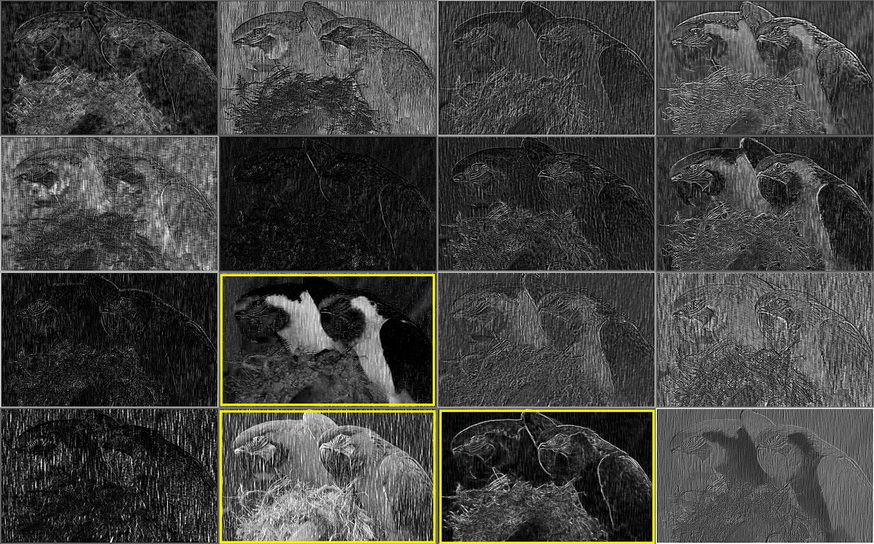}}
\subfigure[Fusing (a) and (b)]{\includegraphics[height = 1in]{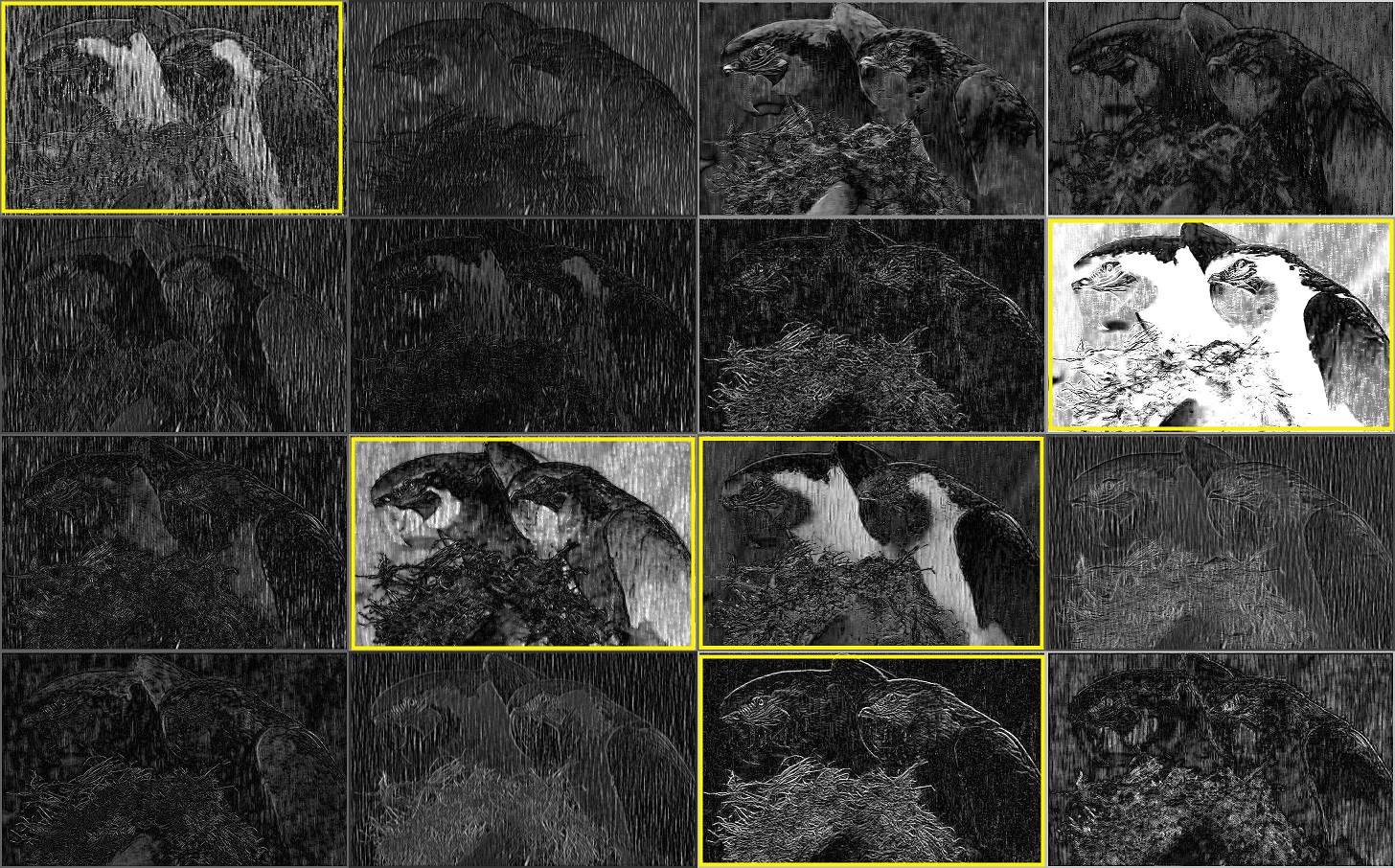}}
\caption{One example of fusion effect \textbf{across} blocks. The fused features are generated by using Equation (\ref{eq.fuse}). Yellow rectangles indicates that the fused features (d) contain more effective representations for \textbf{content} information, which have relatively high contrast to recognize objects.}
\label{fig.blocks}
\end{center}
\end{figure}
The tree-structured fusion is also used across blocks, which we illustrate in Figure \ref{fig.blocks} for the third and forth blocks. As can be seen in Figures \ref{fig.blocks}(a) and (b), the two feature maps have similar content, and so features from adjacent blocks are still redundant. As shown in Figure \ref{fig.blocks}(d), compared with the two input features, the fused feature maps not only remain similar in their high-frequency content, but also generate new representations. Moreover, compared with direct addition, shown in Figures \ref{fig.dilated}(c) and \ref{fig.blocks}(c), using the proposed hierarchical fusion can generate more effective spatial and content representations.

\begin{figure}
\centering
\subfigure[Within the 1st block]{\includegraphics[width = 0.49\columnwidth]{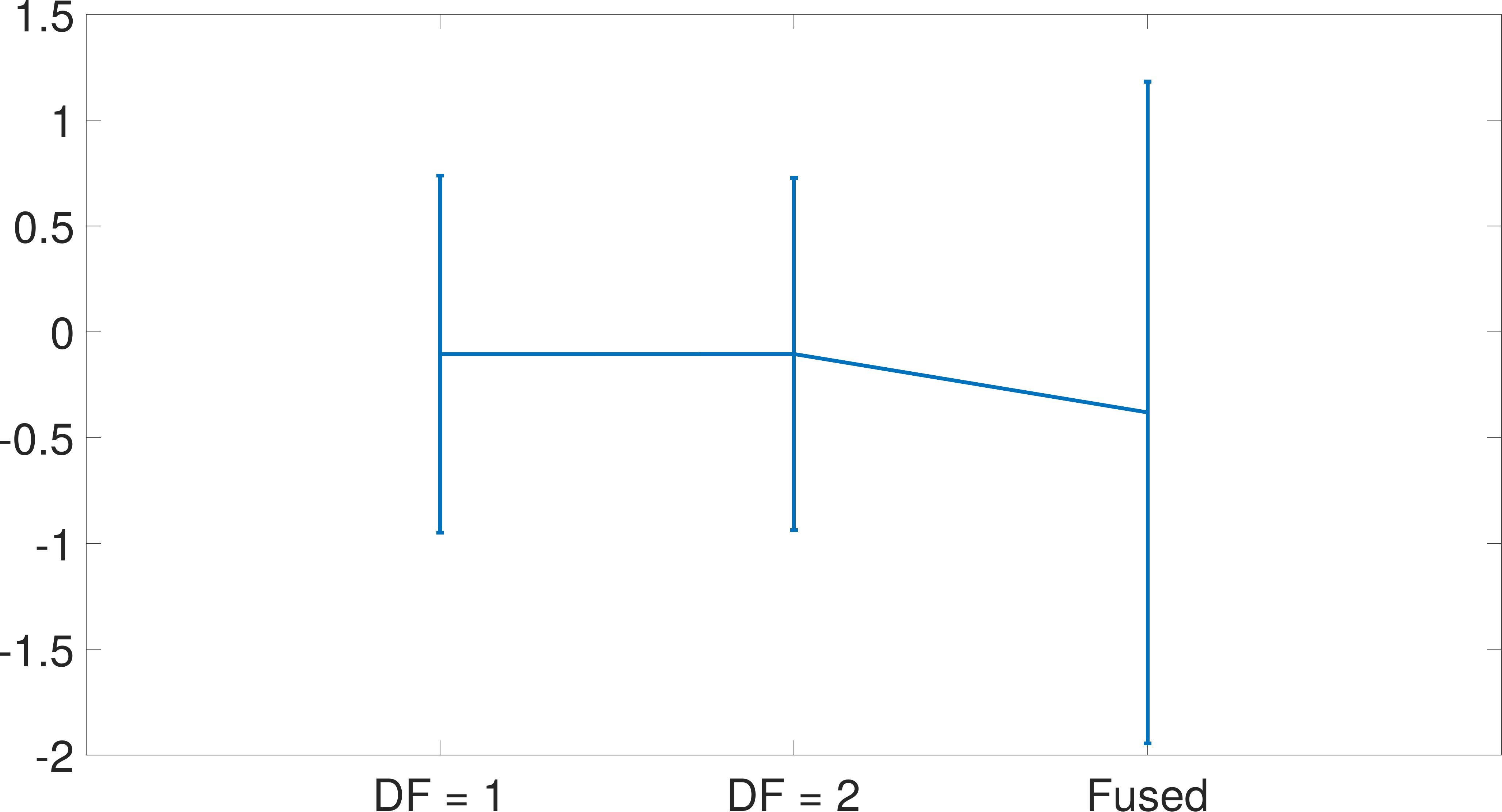}}
\subfigure[Within the 3rd block]{\includegraphics[width = 0.49\columnwidth]{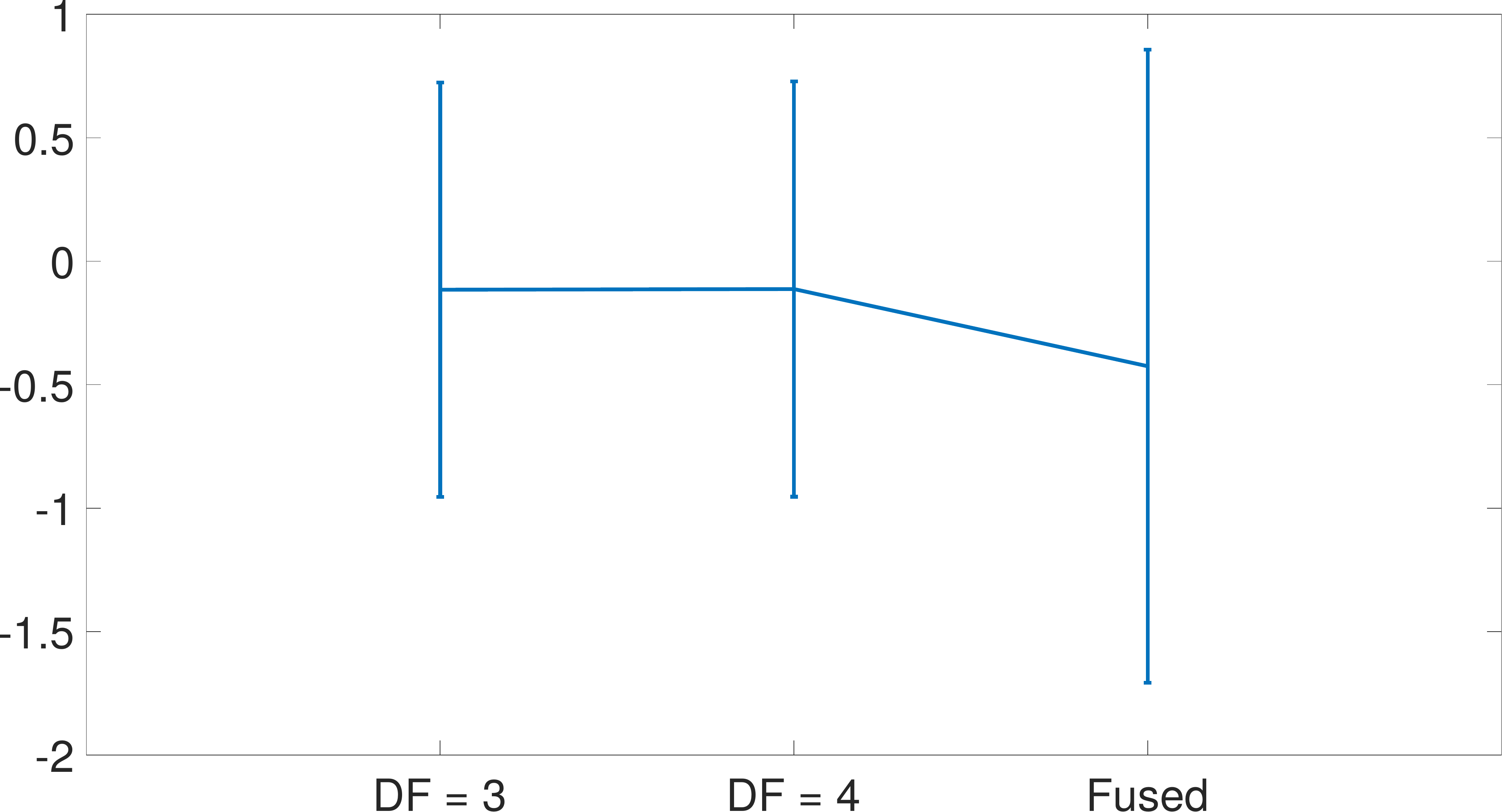}}
\subfigure[Across 1st and 2nd blocks]{\includegraphics[width = 0.49\columnwidth]{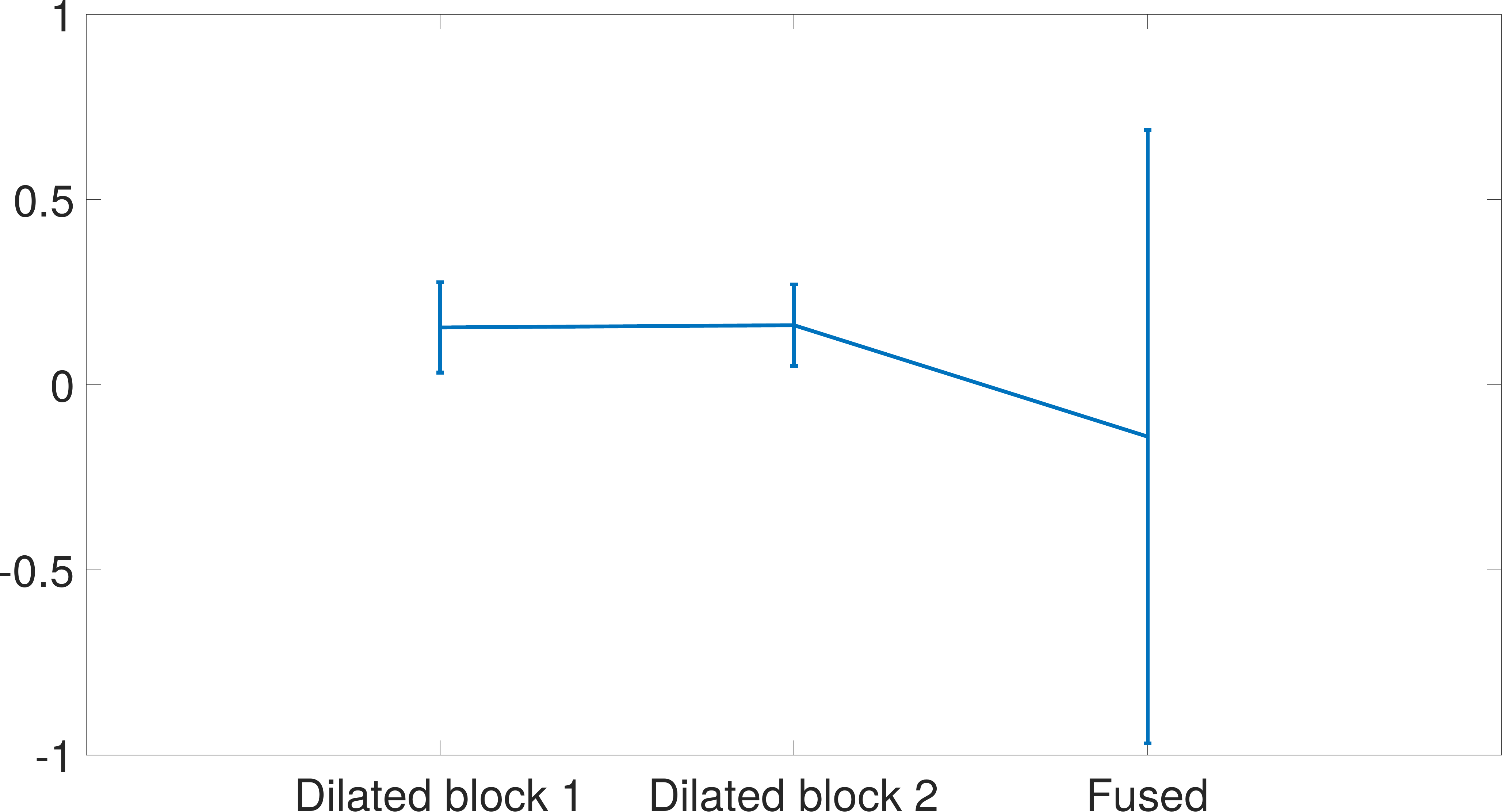}}
\subfigure[Across 5th and 6th blocks]{\includegraphics[width = 0.49\columnwidth]{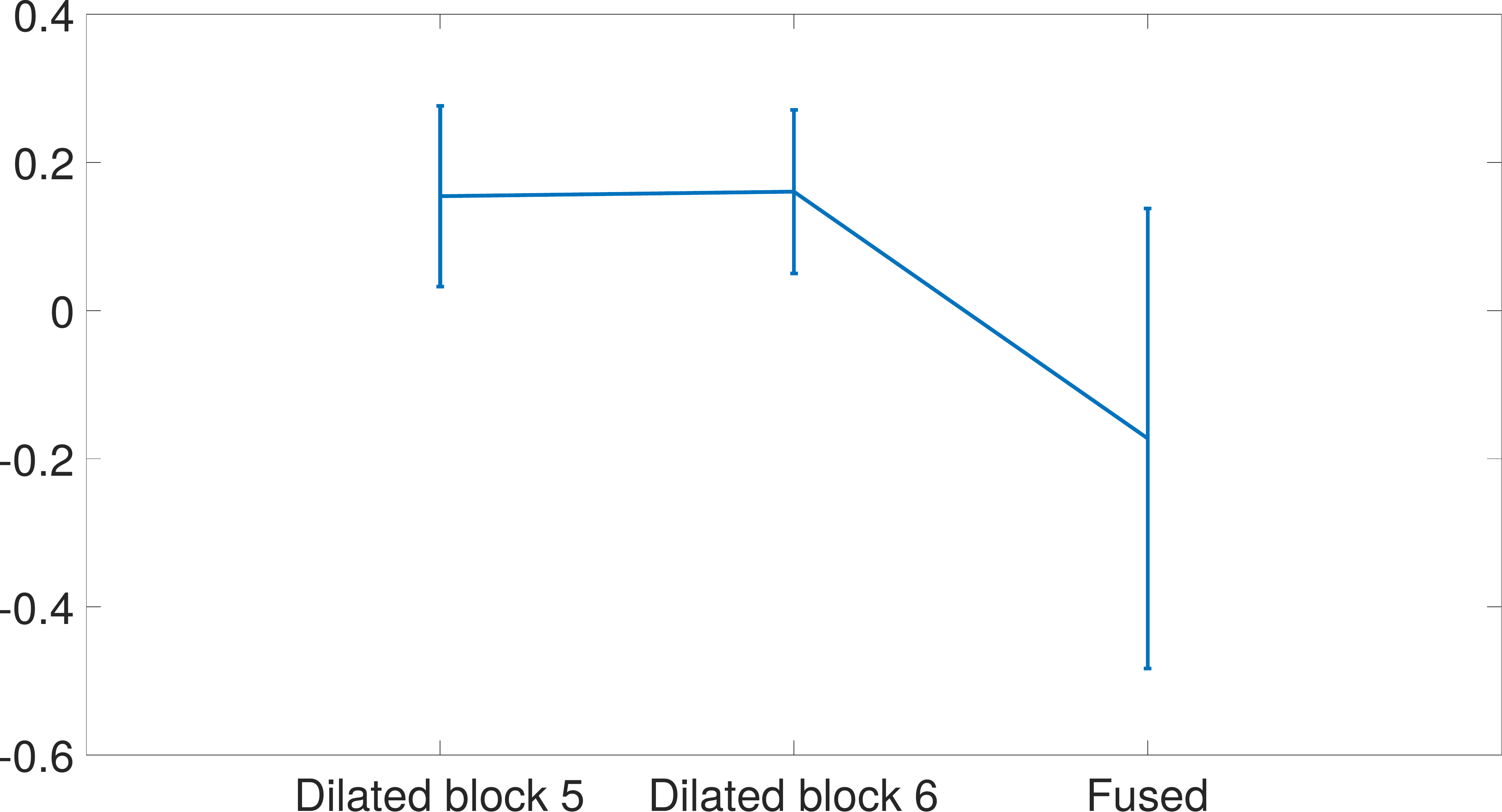}}
\caption{Error bars verifying the redundancy in adjacent features.}
\label{fig.errorbars}
\end{figure}

We show a statistical analysis of this redundancy in Figure \ref{fig.errorbars}. Figures \ref{fig.errorbars}(a) and (b) show statistics of the difference between adjacent features generated by different dilation factors. It is clear that adjacent features are similar, indicating a duplication of information, as also illustrated in Figures \ref{fig.dilated}(a) and (b). This redundancy also exists across blocks, as shown in Figures \ref{fig.errorbars}(c) and (d). This is because for this regression task the resolution of feature maps at deeper layers are the same as the input image , meaning deeper features have no significant change in global content \cite{fu2017removing,Yang2017Deep,Li2018Recurrent}. This is in contrast to high-level vision problems that require pooling operations to extract high-level semantic information \cite{krizhevsky2012imagenet,he2016deep}. To a certain extent the redundancy in global content will persist as the network deepens, motivating fusion in this direction as well. The corresponding fused features have a significant change, shown in Figure \ref{fig.errorbars}. The average appears shifted and the standard deviation becomes larger, indicating that the fused features remove redundant spatial information, also illustrated in Figures \ref{fig.dilated}(d) and \ref{fig.blocks}(d).

\subsection{Loss Function}
The most widely used loss function for training a network is mean squared error. However, MSE usually generates over-smoothed results because of its $\ell_2$ penalty. To address this drawback, we combine the MSE with the SSIM \cite{Wang2014SSIM} as our loss function to balance between rain removal and detail preservation,
\begin{align}
\label{eq.loss}
\mathcal{L}= &\frac{1}{M}\sum\limits_{i = 1}^M {{\alpha  \mathcal{L}_M({{\bf{Y}}_i},{\widehat{\bf{Y}}}_i})} + {(1-\alpha)  {\mathcal{L}_S({{\bf{Y}}_i},{\widehat{\bf{Y}}}_i} )},
\end{align}
where $\mathcal{L}_M$ and $\mathcal{L}_S$ are the MSE and SSIM loss, respectively. $M$ is the number of training data and $\widehat{\bf{Y}}$ is the ground truth. $\alpha$ is the parameter to balance the two losses.

\subsection{Parameters settings and training details}
We set the size of the kernels in Equations (\ref{eq.eq4}) and (\ref{eq.fuse}) are $1 \times 1$ and the rest are $3 \times 3$. The number of feature maps is $16$ for all convolutions and the non-linear activation $\sigma(\cdot)$ is ReLU \cite{krizhevsky2012imagenet}. The dilated factor is set from $1$ to $4$ within each block. We found $8$ dilated convolution blocks are enough to generate good results. We set $\alpha=0.4$. The layer number $L=10$ which indicates two convolutional layers and eight blocks shown in Figure \ref{fig.framework}. All network components can be built by using standard CNN techniques. Other activation functions and network structures can be directly embedded, such as xUnit \cite{Kligvasser2018CVPR} and recurrent architectures \cite{Li2018Recurrent}.

We use TensorFlow \cite{abadi2016tensorflow} and Adam \cite{Kingma2014Adam} with a mini-batch size of 10 to train our network. We initialize the learning rate to 0.001, divide it by 10 at 100K and 200K iterations, and terminate training after 300K iterations. We randomly select $100 \times 100$ patch pairs from training image datasets as inputs. All experiment are performed on a server with Intel(R) Xeon(R) CPU E5-2683, 64GB RAM and NVIDIA GTX 1080. The network is trained end-to-end.

\section{Experiments}
We compare our network with two model-based deraining methods: the Gaussian Mixture Model (GMM) \cite{Li2016Rain} and Joint Convolutional Analysis and Synthesis (JCAS) \cite{gu2017joint}. We also compare with four deep learning-based methods: Deep Detail Network (DDN) \cite{fu2017removing}, JOint Rain DEtection and Removal (JORDER) \cite{Yang2017Deep}, Density-aware Image Deraining (DID) \cite{zhang2018density} and REcurrent Squeeze-and-excitation Context Aggregation Net (RESCAN) \cite{Li2018Recurrent}. All methods are retrained for a fair and meaningful comparison.\footnote{Due to retraining and different data, the quantitative results may be different from the results reported in the corresponding articles.}

\begin{table*}
\caption{Average SSIM and PSNR values on synthesized images. The best and the second best results are boldfaced and underlined. Numbers in parentheses indicates the parameter reduction.}
\resizebox{\linewidth}{!}{
\begin{tabular}{|*{15}{c|}}
\hline
& \multicolumn{2}{|c|}{GMM} & \multicolumn{2}{|c|}{JCAS}& \multicolumn{2}{|c|}{DDN} & \multicolumn{2}{|c|}{JORDER}& \multicolumn{2}{|c|}{DID}
& \multicolumn{2}{|c|}{RESCAN} & \multicolumn{2}{|c|}{Ours} \\ \cline{1-15}
Datasets         & SSIM &PSNR  &SSIM  &PSNR   &SSIM  &PSNR              &SSIM  &PSNR  &SSIM            &PSNR  &SSIM             &PSNR              &SSIM           &PSNR\\
\hline
\emph{Rain100H}  &0.43  &15.05 &0.51  &15.23  &0.81  &\underline{26.88} &0.84  &26.54 &0.83            &26.12 &\underline{0.85} &26.45             &\textbf{0.88}  &\textbf{27.46}  \\
\hline
\emph{Rain1400}  &0.83  &26.53 &0.85  &26.80  &0.89  &29.99             &0.90  &28.90 &0.90            &29.84 &\underline{0.91} &\underline{31.18} &\textbf{0.92}  &\textbf{31.32}   \\
\hline
\emph{Rain1200}  &0.80  &22.46 &0.81  &25.16  &0.86  &30.95             &0.87  &29.75 &\underline{0.90}&29.65 &0.89             &\textbf{32.35}    &\textbf{0.92}  &\underline{32.30} \\
\hline
Parameters \# & \multicolumn{2}{|c|}{-} & \multicolumn{2}{|c|}{-}&\multicolumn{2}{|c|}{58,175 (-39\%)} & \multicolumn{2}{|c|}{369,792 (-90\%)} & \multicolumn{2}{|c|}{372,839 (-90\%)}&  \multicolumn{2}{|c|}{54,735 (-35\%)} & \multicolumn{2}{|c|}{\textbf{35,427}} \\ \cline{1-15}
\end{tabular}
}
\label{tab.SSIM}
\end{table*}

\begin{table*}
\caption{Comparison of running time (seconds).}
\resizebox{\linewidth}{!}{
\begin{tabular}{|*{13}{c|}}
\hline
& GMM & JCAS & \multicolumn{2}{|c|}{DDN} & \multicolumn{2}{|c|}{JORDER}&  \multicolumn{2}{|c|}{DID}& \multicolumn{2}{|c|}{RESCAN}& \multicolumn{2}{|c|}{Ours}  \\\cline{1-13}
Image size         &CPU& CPU  & CPU & GPU & CPU  & GPU  & CPU  &GPU & CPU  &GPU& CPU  &GPU   \\\hline
512 $\times$ 512   &1.99$\times$10$^3$&0.97$\times$10$^2$  &1.51 &0.16 &2.95$\times$10$^2$  & 0.18 &7.26               &0.31 &6.29               &0.13&1.94  &0.16 \\\hline
1024 $\times$ 1024 &6.52$\times$10$^3$&6.57$\times$10$^2$  &5.40 &0.32 &1.20$\times$10$^3$  & 0.82 &1.78$\times$10$^2$ &0.78 &1.23$\times$10$^2$ &0.28&7.51  &0.28  \\\hline
\end{tabular}
}
\label{tab:time}
\end{table*}

\subsection{Synthetic data}
We us three public synthetic datasets provided by JORDER \cite{Yang2017Deep}, DDN \cite{fu2017removing} and DID \cite{zhang2018density}. These three datasets were generated using different synthetic strategies. The JORDER dataset contains 100 testing images with heavy rain streaks. The other two contain 1400 and 1200 testing images, respectively. We call them \emph{Rain100H}, \emph{Rain1400} and \emph{Rain1200} below.
\begin{figure}[t]
\begin{center}
\subfigure[Clean $|$ SSIM]{\includegraphics[width = 1.05in]{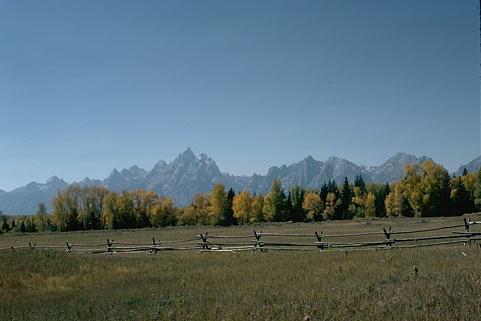}}
\subfigure[Input  $|$  0.81]{\includegraphics[width = 1.05in]{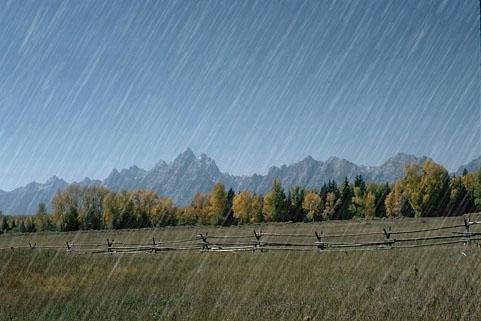}}
\subfigure[GMM  $|$ 0.90]{\includegraphics[width = 1.05in]{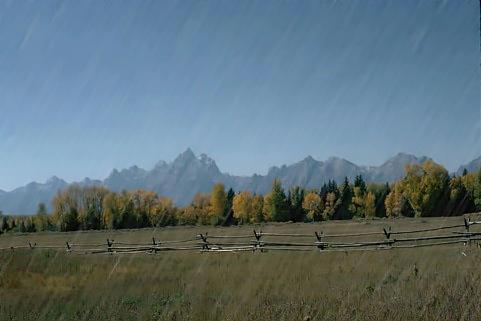}}\\
\subfigure[JCAS  $|$  0.90]{\includegraphics[width = 1.05in]{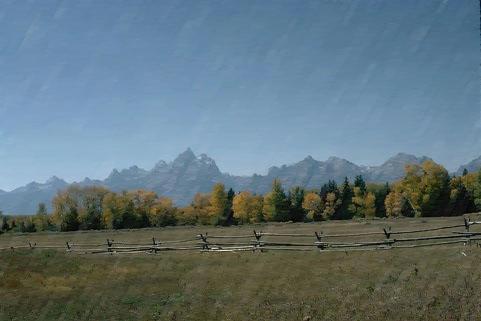}}
\subfigure[DDN  $|$ 0.89]{\includegraphics[width = 1.05in]{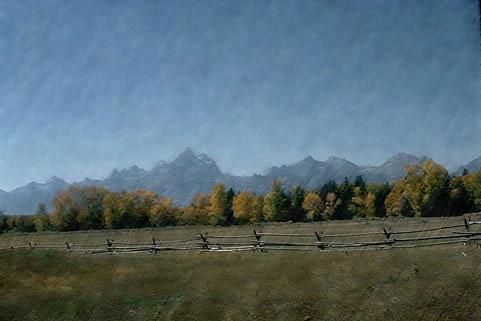}}
\subfigure[JORDER  $|$  0.91]{\includegraphics[width = 1.05in]{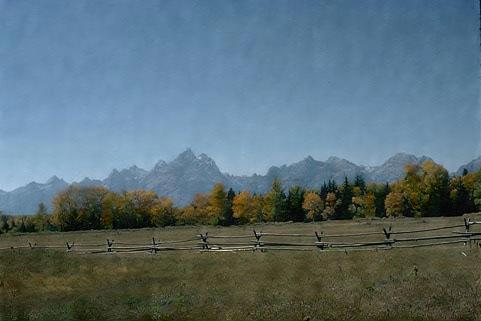}}\\
\subfigure[DID  $|$ 0.91]{\includegraphics[width = 1.05in]{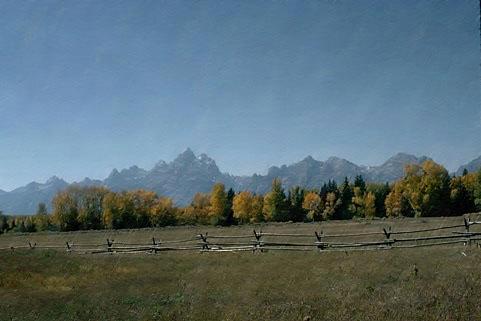}}
\subfigure[RESCAN  $|$  \textbf{0.93}]{\includegraphics[width = 1.05in]{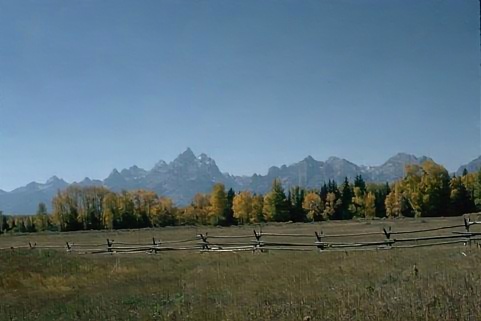}}
\subfigure[Ours  $|$  \textbf{0.93}]{\includegraphics[width = 1.05in]{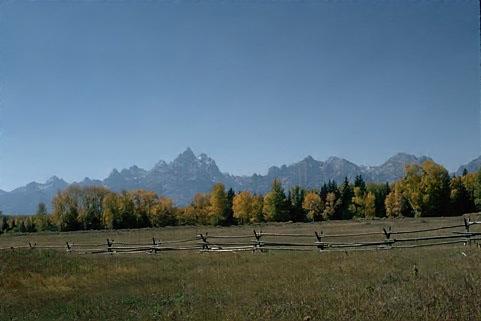}}
\caption{One visual comparisons on `\emph{Rain1400}' dataset.}
\label{fig.synthetic1}
\vspace{0.1in}
\subfigure[Clean  $|$  SSIM]{\includegraphics[width = 1.05in]{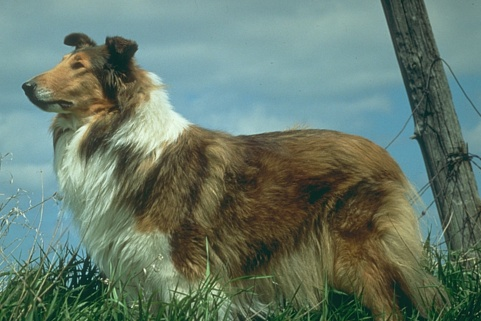}}
\subfigure[Input  $|$  0.35]{\includegraphics[width = 1.05in]{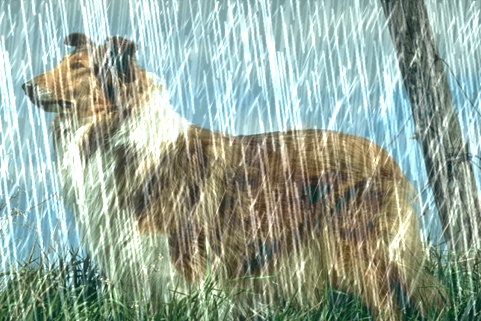}}
\subfigure[GMM  $|$ 0.41]{\includegraphics[width = 1.05in]{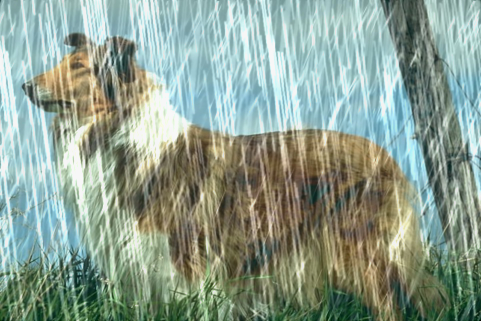}}\\
\subfigure[JCAS  $|$  0.49]{\includegraphics[width = 1.05in]{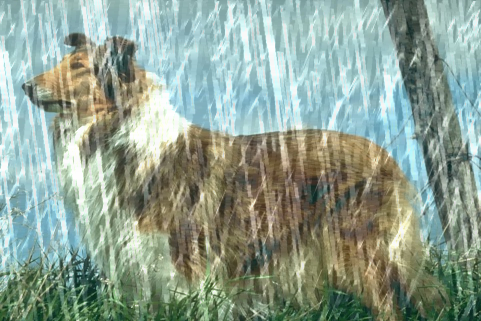}}
\subfigure[DDN   $|$  0.78]{\includegraphics[width = 1.05in]{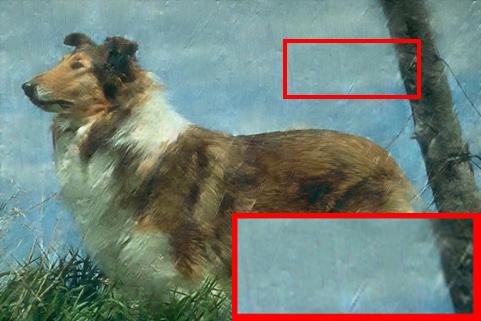}}
\subfigure[JORDER  $|$  0.81]{\includegraphics[width = 1.05in]{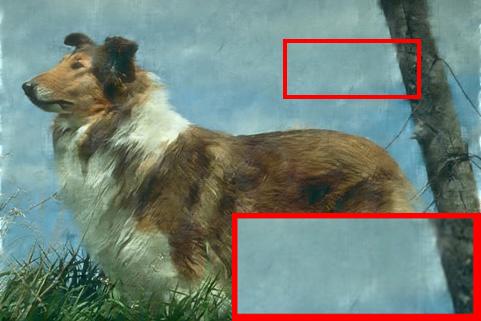}}\\
\subfigure[DID $|$  0.82]{\includegraphics[width = 1.05in]{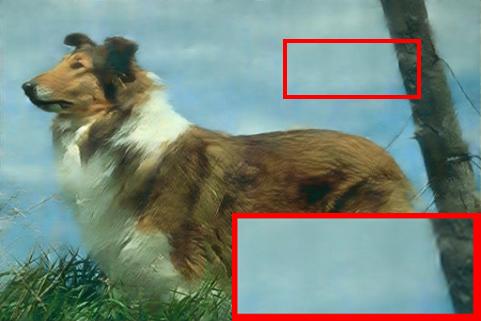}}
\subfigure[RESCAN  $|$  0.79]{\includegraphics[width = 1.05in]{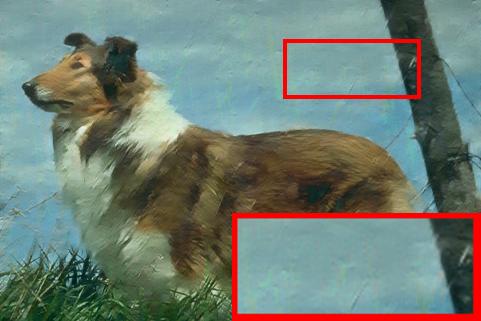}}
\subfigure[Ours  $|$ \textbf{0.85}]{\includegraphics[width = 1.05in]{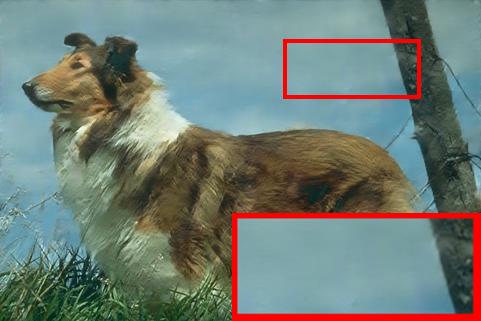}}
\caption{One visual comparisons on `\emph{Rain100H}' dataset.}
\label{fig.synthetic2}
\end{center}
\end{figure}
\begin{figure}[t]
\begin{center}
\subfigure[Clean  $|$  SSIM]{\includegraphics[width = 1.05in]{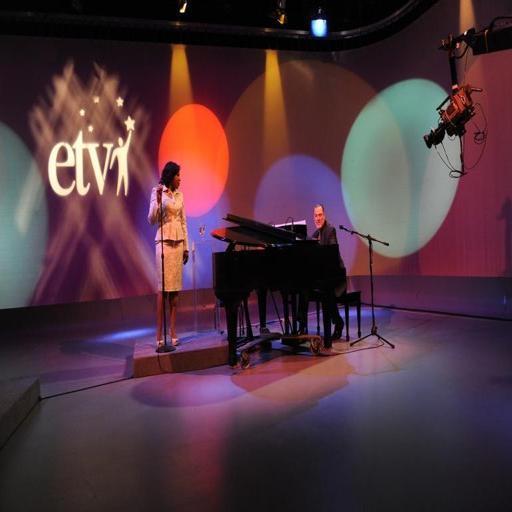}}
\subfigure[Input  $|$ 0.29]{\includegraphics[width = 1.05in]{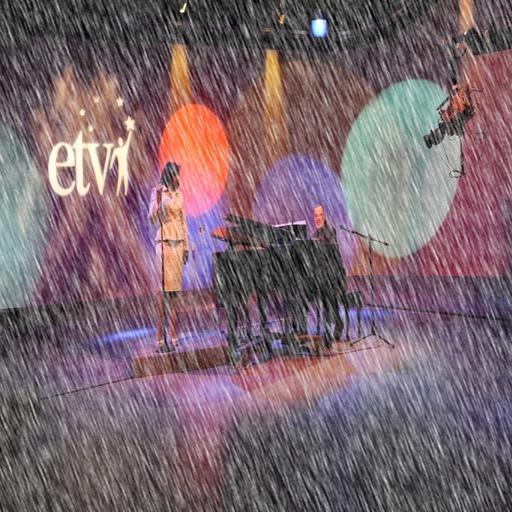}}
\subfigure[GMM  $|$ 0.45]{\includegraphics[width = 1.05in]{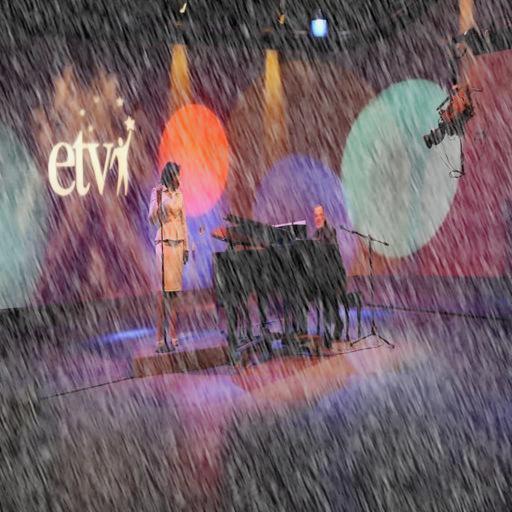}}\\
\subfigure[JCAS  $|$ 0.51]{\includegraphics[width = 1.05in]{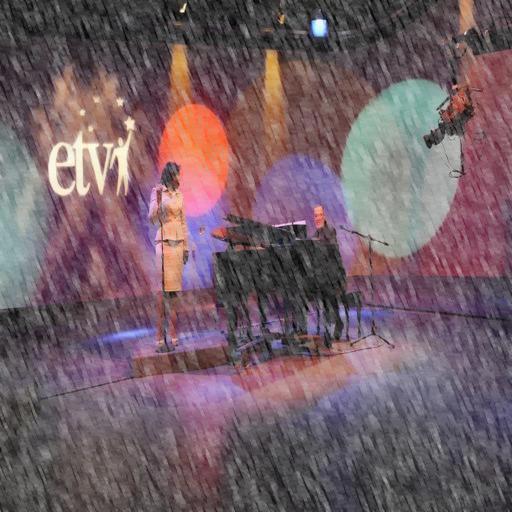}}
\subfigure[DDN  $|$ 0.82]{\includegraphics[width = 1.05in]{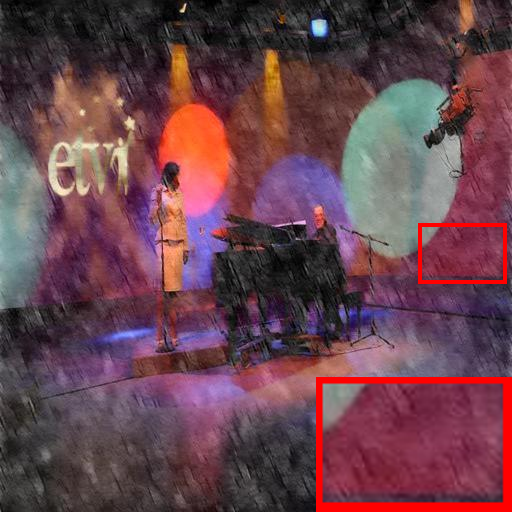}}
\subfigure[JORDER  $|$ 0.86]{\includegraphics[width = 1.05in]{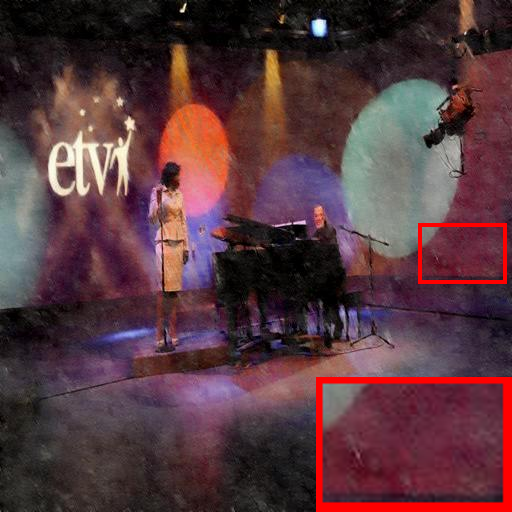}}\\
\subfigure[DID  $|$ \textbf{0.91}]{\includegraphics[width = 1.05in]{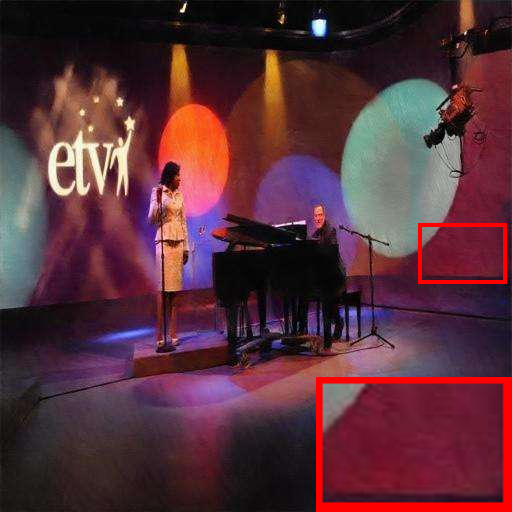}}
\subfigure[RESCAN  $|$  0.89]{\includegraphics[width = 1.05in]{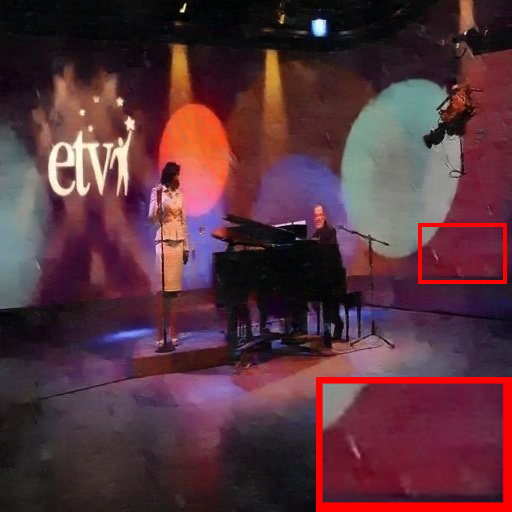}}
\subfigure[Ours  $|$ \textbf{0.91}]{\includegraphics[width = 1.05in]{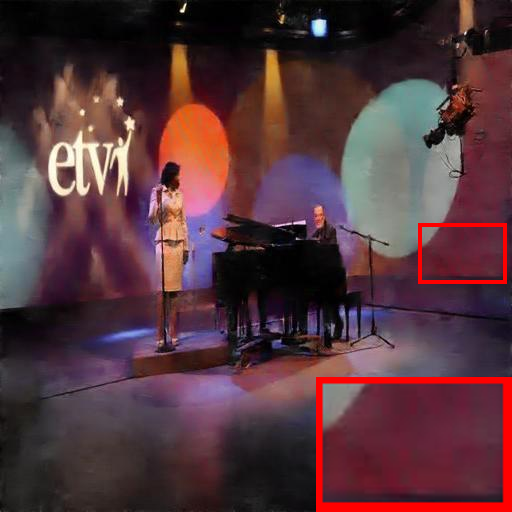}}
\caption{One visual comparisons on `\emph{Rain1200}' dataset.}
\label{fig.synthetic3}
\end{center}
\end{figure}

Figures \ref{fig.synthetic1} to \ref{fig.synthetic3} show three visual results from each dataset with different rain orientations and magnitudes. It is clear that GMM and JCAS fail due to modeling limitations. As shown in the red rectangles, DDN and JORDER are able to remove the rain streaks while tending to generate artifacts. DID has a good deraining performance while slightly blurring edges, shown in Figure \ref{fig.synthetic3}(e) and (g). RESCAN and our model have similar global visual performance and outperform other methods. We also calculate PSNR and SSIM for quantitative evaluation in Table \ref{tab.SSIM}. Our method has the best overall results on both PSNR and SSIM, indicating our tree-structured fusion can better represent spatial information. Moreover, our network contains the fewest parameters which makes it more suitable for practical applications.

\subsection{Real-world data}
\begin{figure*}[t]
\begin{center}
\subfigure[Input]{\includegraphics[width = .24\textwidth]{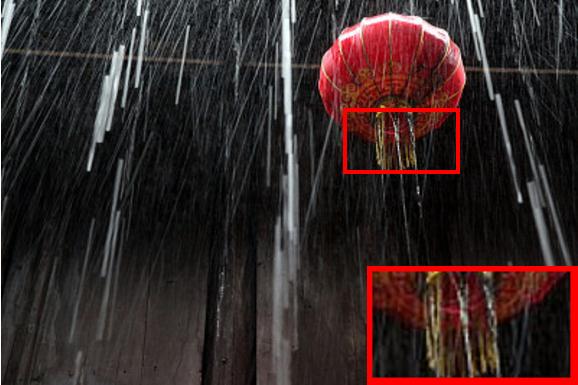}}
\subfigure[GMM]{\includegraphics[width = .24\textwidth]{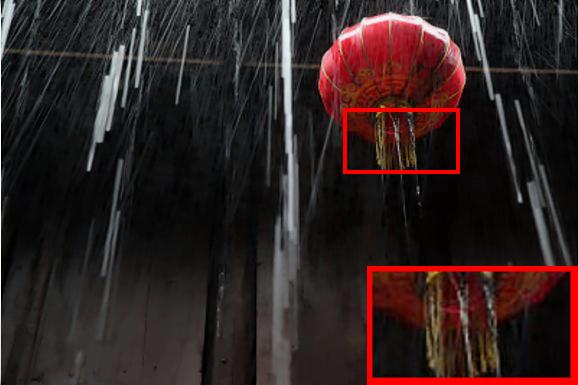}}
\subfigure[JCAS]{\includegraphics[width = .24\textwidth]{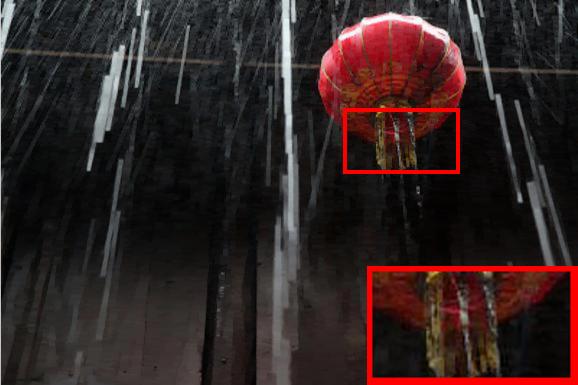}}
\subfigure[DDN]{\includegraphics[width = .24\textwidth]{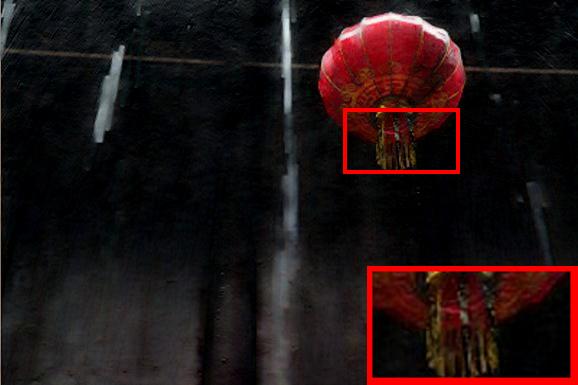}}\\
\subfigure[JORDER]{\includegraphics[width = .24\textwidth]{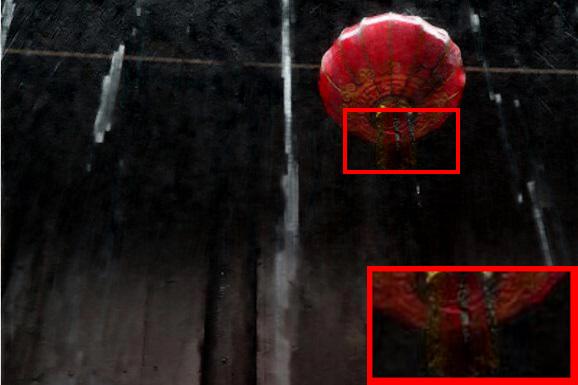}}
\subfigure[DID]{\includegraphics[width = .24\textwidth]{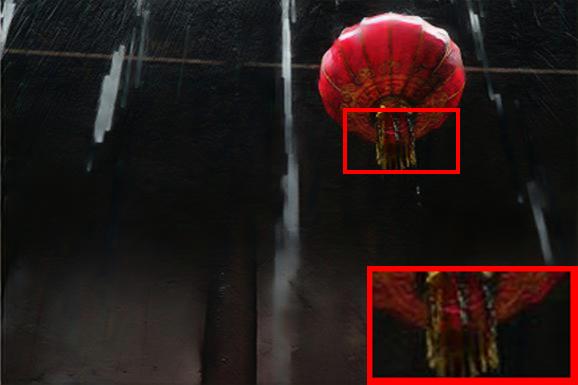}}
\subfigure[RESCAN]{\includegraphics[width = .24\textwidth]{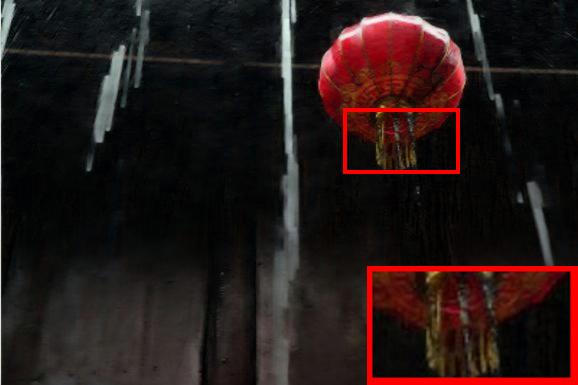}}
\subfigure[Our]{\includegraphics[width = .24\textwidth]{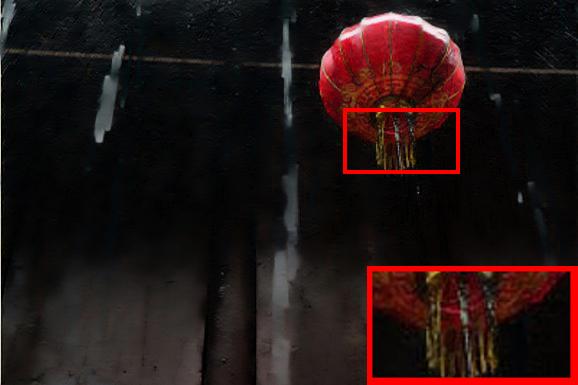}}
\caption{Visual comparisons on real-world rainy images.}
\label{fig.real1}
\vspace{0.1in}
\subfigure[Input]{\includegraphics[width = .24\textwidth]{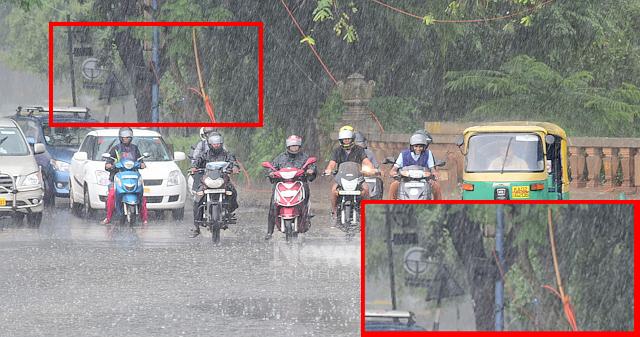}}
\subfigure[GMM]{\includegraphics[width = .24\textwidth]{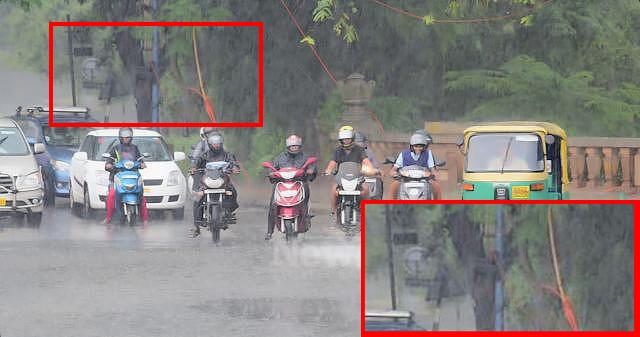}}
\subfigure[JCAS]{\includegraphics[width = .24\textwidth]{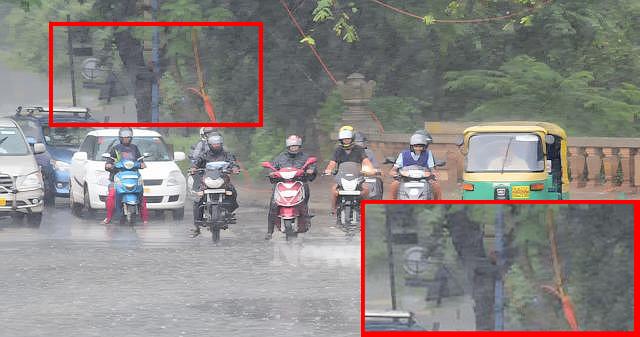}}
\subfigure[DDN]{\includegraphics[width = .24\textwidth]{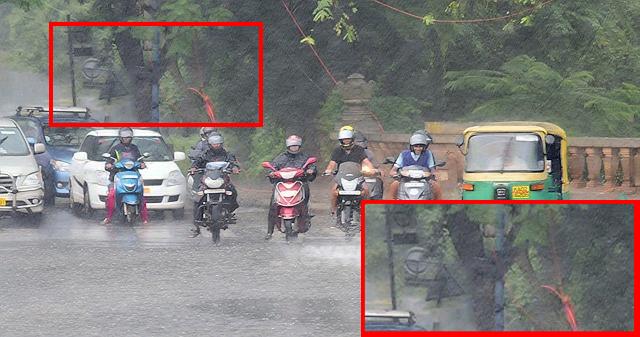}}\\
\subfigure[JORDER]{\includegraphics[width = .24\textwidth]{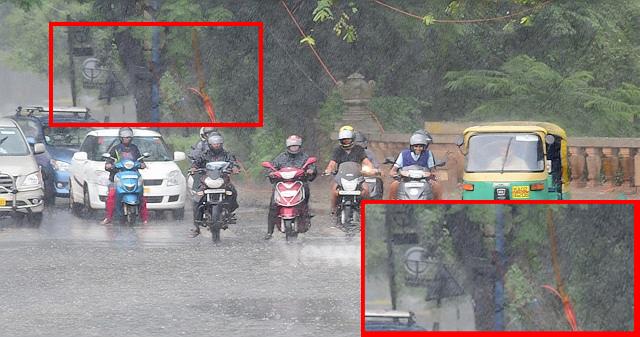}}
\subfigure[DID]{\includegraphics[width = .24\textwidth]{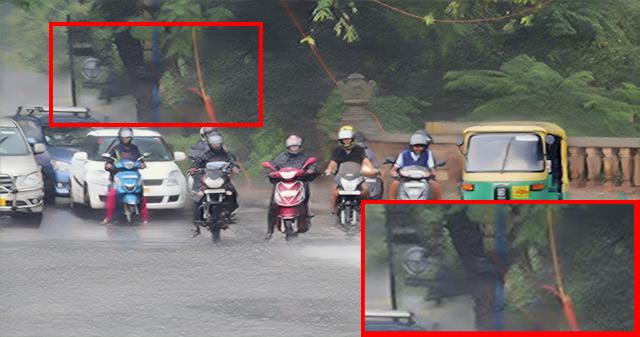}}
\subfigure[RESCAN]{\includegraphics[width = .24\textwidth]{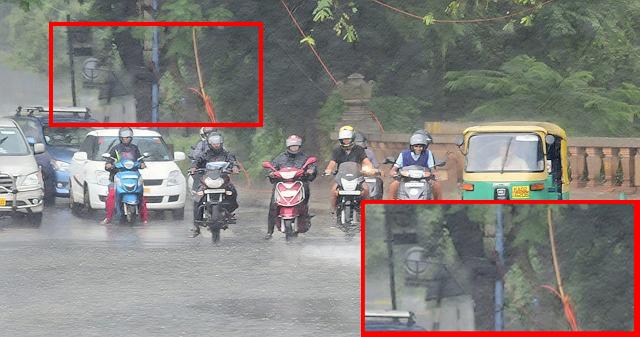}}
\subfigure[Our]{\includegraphics[width = .24\textwidth]{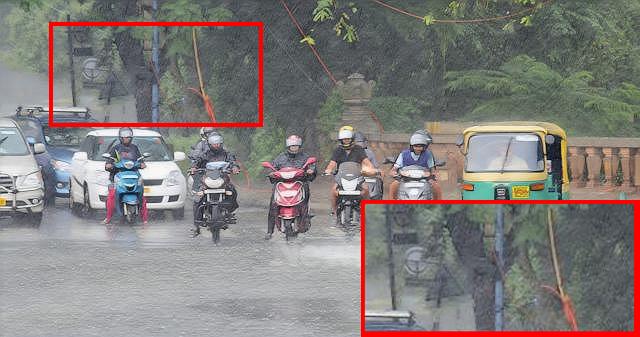}}
\caption{Visual comparisons on real-world rainy images.}
\label{fig.real2}
\end{center}
\end{figure*}

We also show that our learned network, which is trained on synthetic data, translates well to real-world rainy images. Figures \ref{fig.real1} and \ref{fig.real2} show two typical visual results on real-world images. The red rectangles indicate that our network can simultaneously remove rain and preserve details.

\begin{figure*}
\begin{center}
\includegraphics[width = 0.95in]{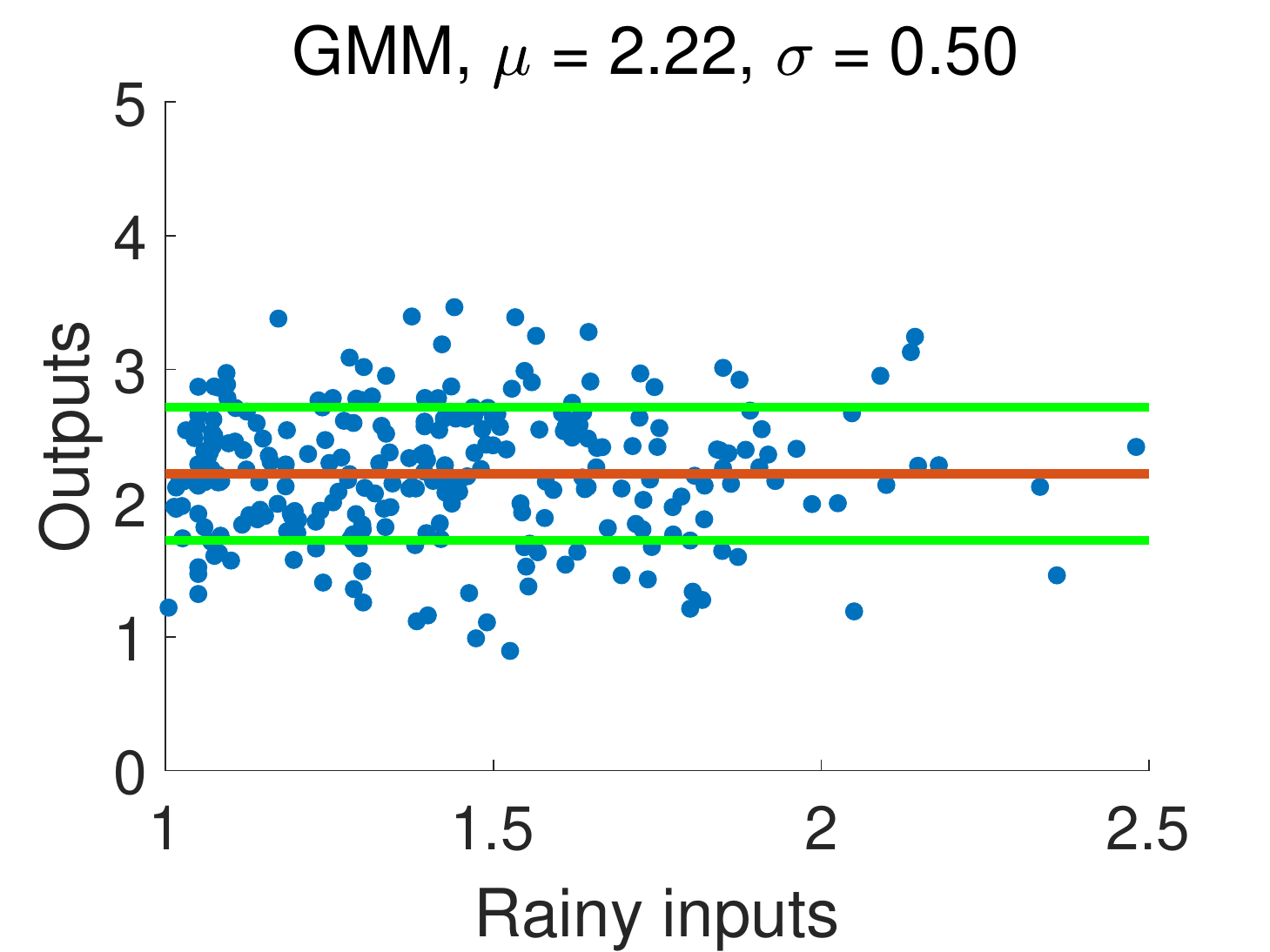}
\includegraphics[width = 0.95in]{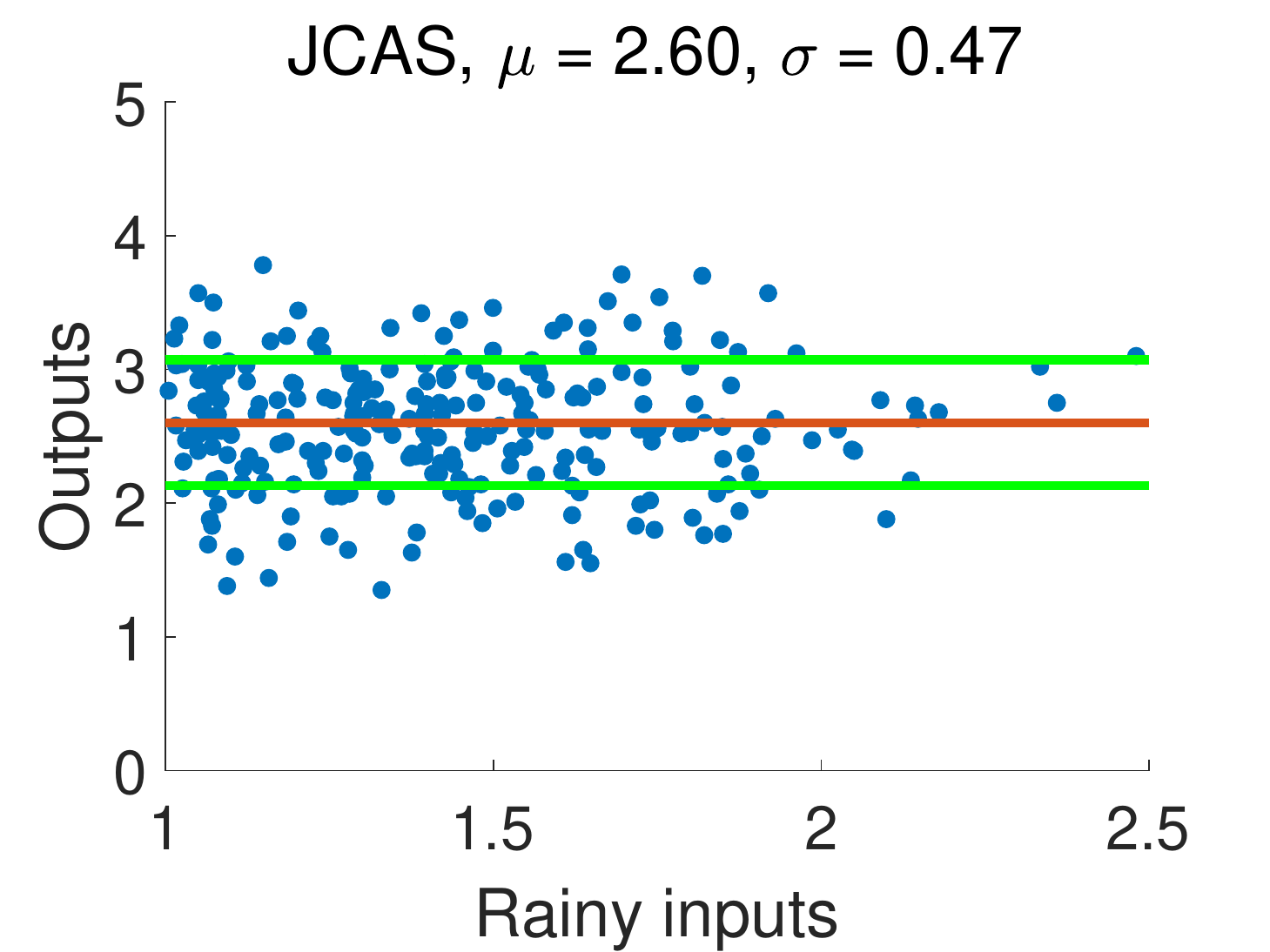}
\includegraphics[width = 0.95in]{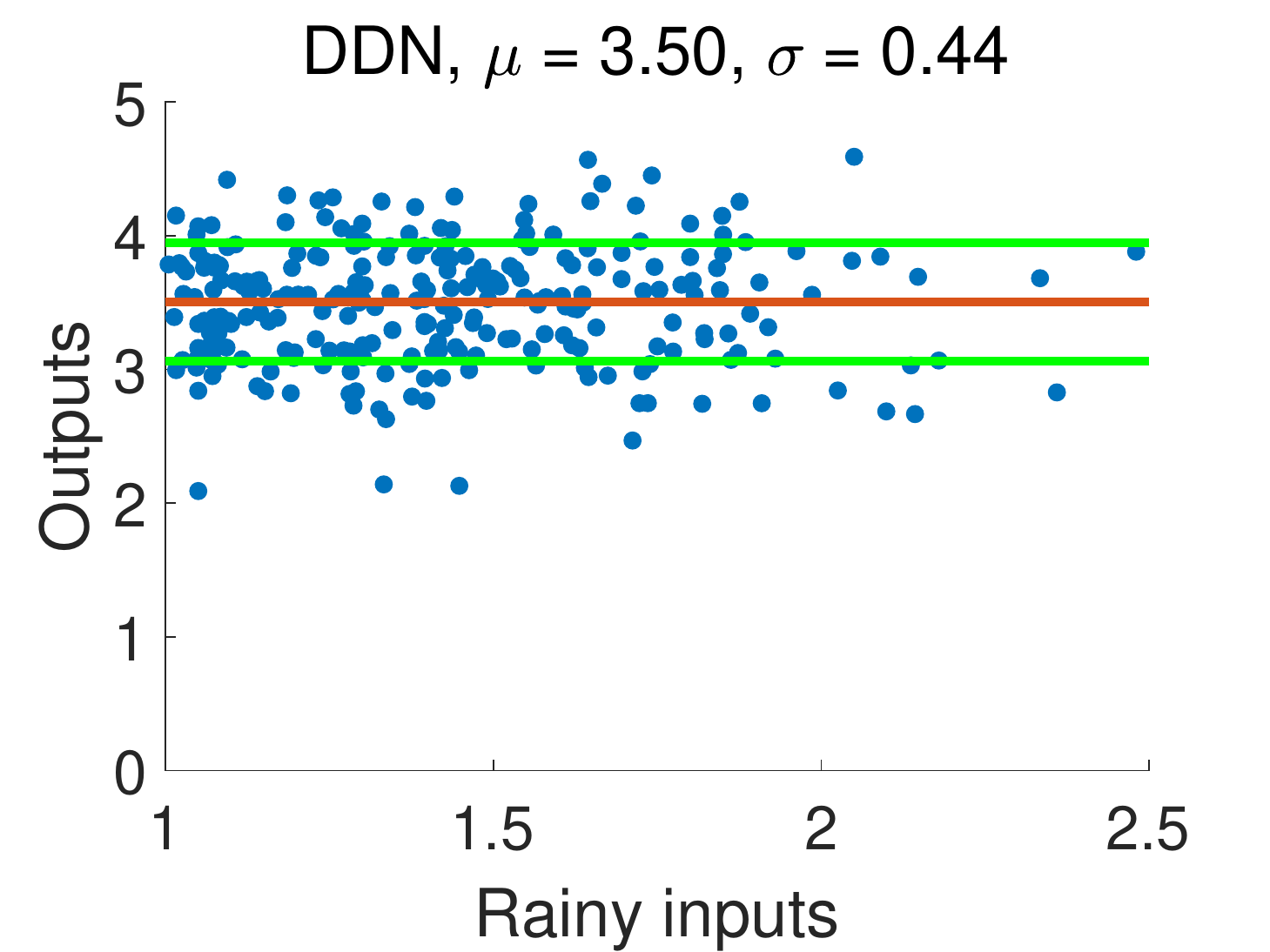}
\includegraphics[width = 0.95in]{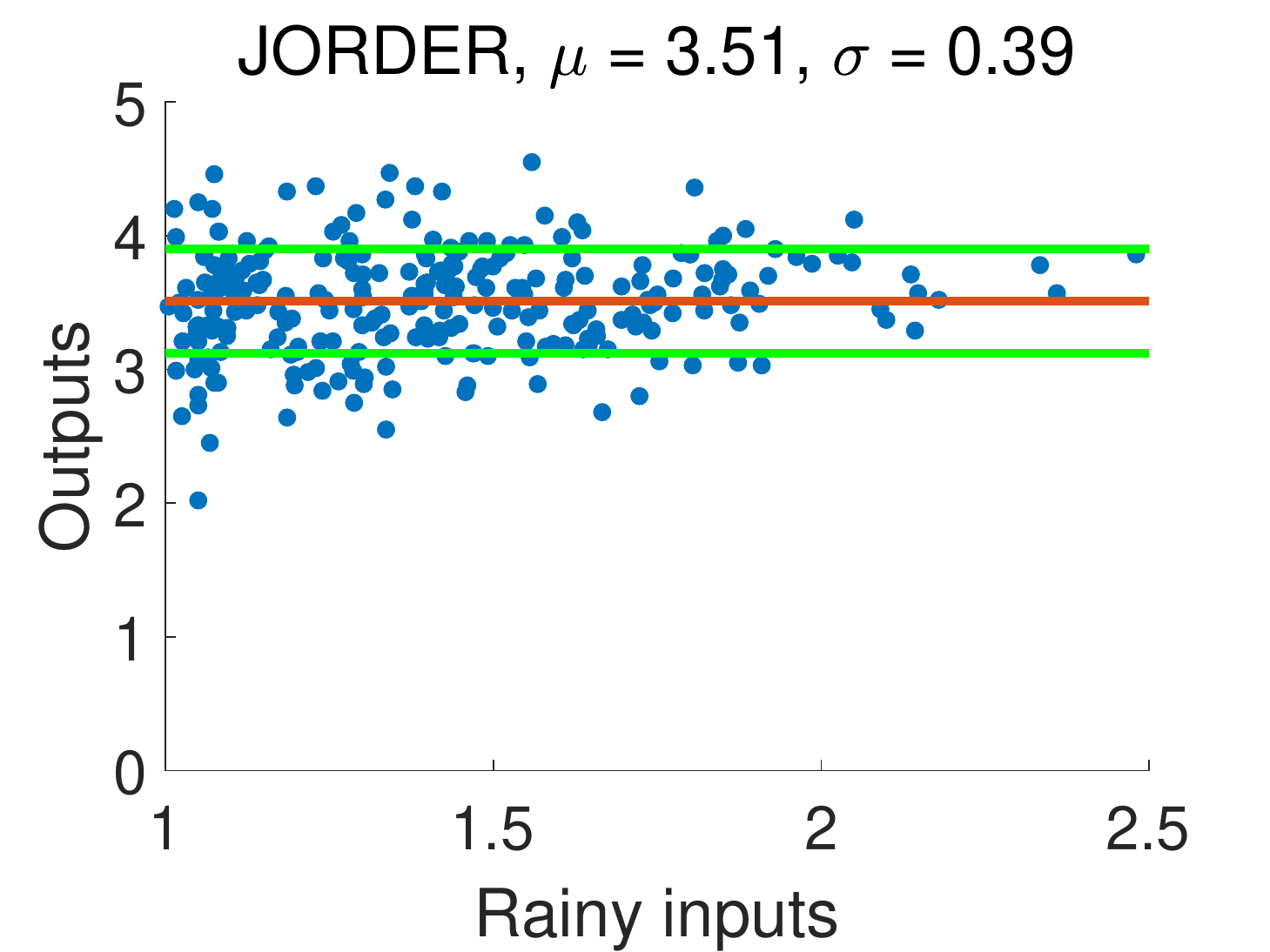}
\includegraphics[width = 0.95in]{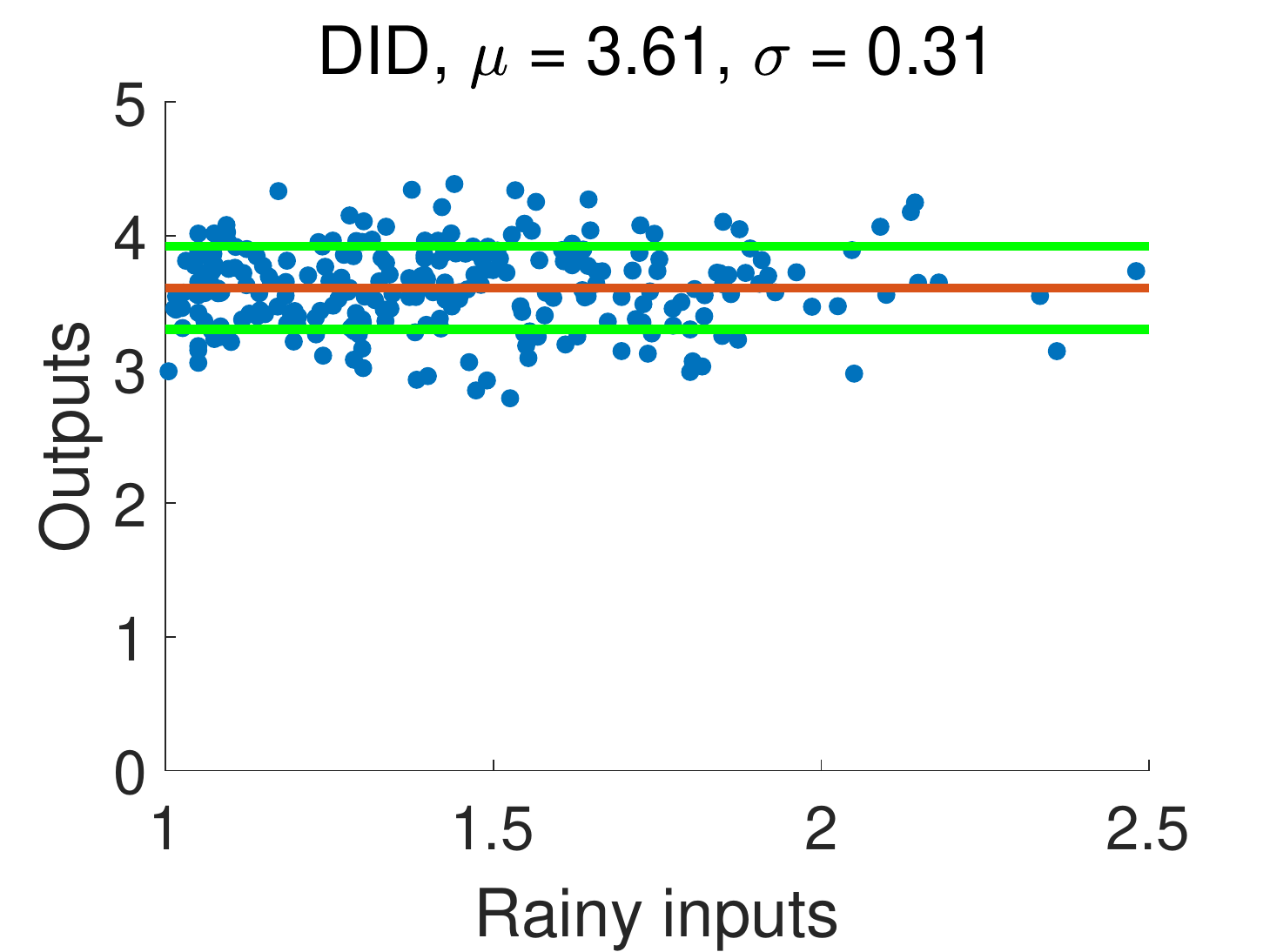}
\includegraphics[width = 0.95in]{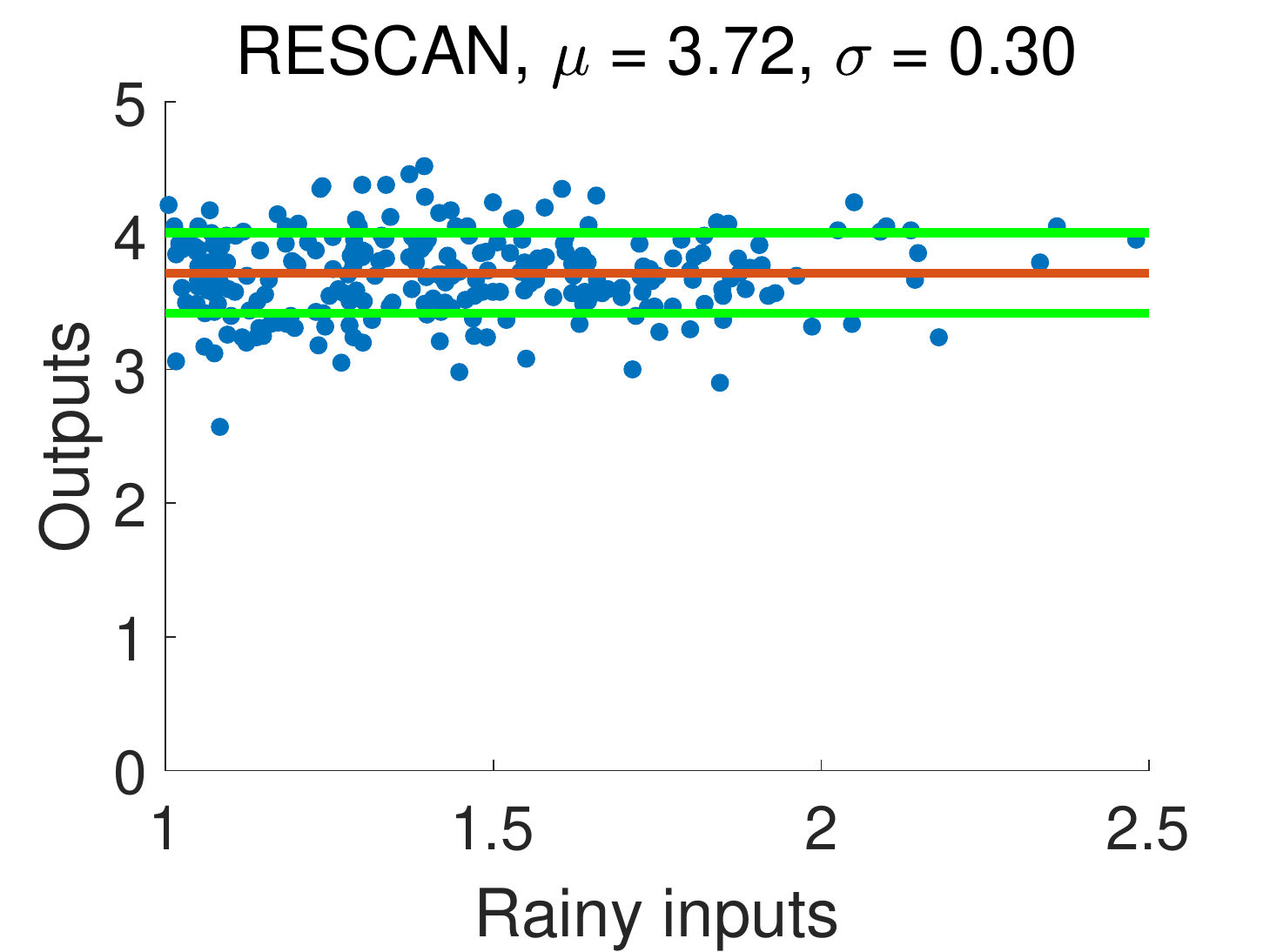}
\includegraphics[width = 0.95in]{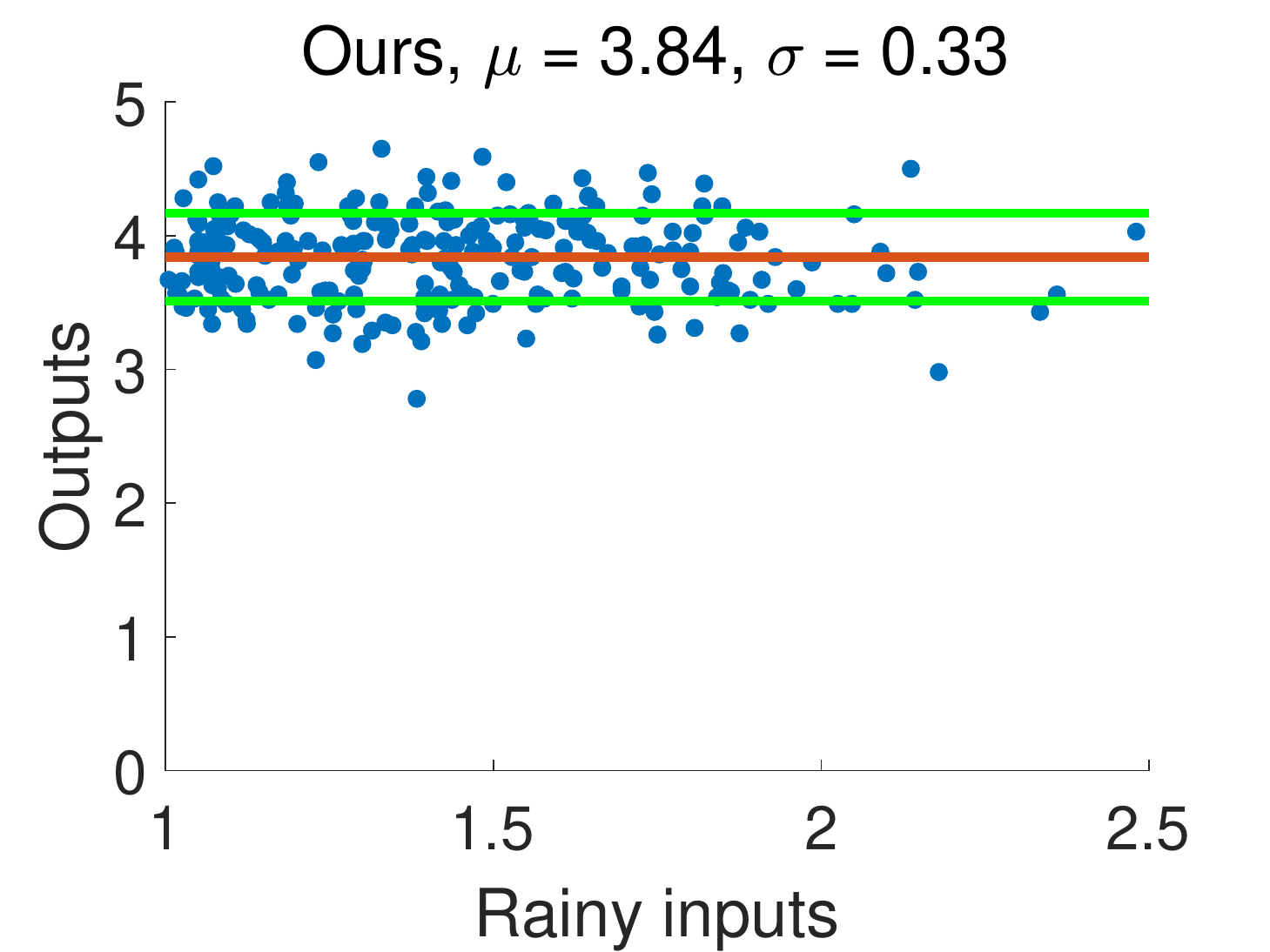}
\caption{Scatter plots of rainy inputs vs derained user scores. Global mean and standard deviation also shown. (Best viewed zoomed in.)}
\label{fig.User}
\end{center}
\end{figure*}

We further collect 300 real-world rainy images from the Internet and existing articles \cite{fu2017removing,Yang2017Deep,zhang2018density} as a new dataset.\footnote{We will release our code and this new dataset.} We then asked $10$ participants to conduct a user study for realistic feedback. The participants are told to rank each derained result from $1$ to $5$ randomly presented without knowing the corresponding algorithm. ($1$ represents the worst quality and $5$ represents the best quality.) We show the scatter plots in Figure \ref{fig.User} where we see our network has the best performance. This provides some additional support for our tree-structured fusion model.

\subsection{Running time}
We show the average running time of different methods in Table \ref{tab:time}. Two different image sizes are chosen and each one is averaged over 100 images. The GMM and JCAS are implemented on CPUs according to the provided code, while other deep learning-based methods are tested on both CPU and GPU. The GMM has the slowest running time since complicated inference is still required to process each new image. Our method has a comparable GPU running time compared with other deep learning methods, and is significantly faster than several deep models on a CPU. This is because our network is straightforward without extra operations, such as recurrent structures \cite{Yang2017Deep,zhang2018density}.

\subsection{Ablation study}
We provide an ablation study to demonstrate the advantage of each part of our network structure.
\subsubsection{Fusion deployment}
To validate our tree-structured feature fusion strategy, we design two variants of the proposed network for exhaustive comparison. One is called Network$^{W}$ that only uses the fusion operation {w}ithin each block. The other is called Network$^{A}$ that only uses the fusion operation {a}cross all blocks. This experiment is trained and tested on the \emph{Rain100H} dataset and we use JORDER as the baseline. Table \ref{tab.Ablation} shows the SSIM performance changes on \emph{Rain100H}. As can be seen, compared with JORDER \cite{Yang2017Deep}, Network$^{A}$ brings a performance increase of 1.56\%, while Network$^{W}$ leads to a 3.35\% improvement. This is because large receptive fields help to capture rain streaks in larger areas, which is a crucial factor for the image deraining problem. However, deploying fusion operations across blocks bring more benefits when building very deep models, since more and richer content information will be generated. By combining the hierarchical structure of Network$^{A}$ and Network$^{W}$ into single final network (our proposed model), the highest SSIM values can be obtained.
\begin{table}[h]
\centering
\caption{Ablation study on fusion deployment.}
\resizebox{\linewidth}{!}{
\label{tab.Ablation}
\begin{tabular}{|c|c|c|c|c|c|c|}
\hline
          & JORDER & Network$^{A} $ &  Network$^{W}$  & Final\\
\hline
SSIM      & 0.835  & 0.848          &   0.863         & 0.877\\
\hline
\end{tabular}
}
\end{table}

\begin{table}[h]
\caption{Ablation study on dilated factors and block numbers.}
\label{tab.blocknumbers}
\resizebox{\linewidth}{!}{
\def\arraystretch{1.25}
\begin{tabular}{|c|c|c|c|c|c|}
\hline
       & $DF =2$ & $DF =4$ (default)   & $DF =6$ \\
\hline
$L=6$  &0.823    &  0.857    & 0.863  \\
\hline
$L=10$ (default)&0.841    &  0.877     & 0.879\\
\hline
$L=14$ &0.846    &  0.881   &  0.882\\
\hline
\end{tabular}
}
\end{table}
\subsubsection{Dilation factor versus block number}
We test the impact of dilation factor and block number on the \emph{Rain100H} dataset. Specifically, we test for dilation factors $DF \in\{2,4,6\}$ and basic block numbers $L\in\{6,10,14\}$. The SSIM results are shown in Table \ref{tab.blocknumbers}. As can be seen, increasing dilation and blocks can generate higher SSIM values. Moreover, larger dilations result in larger receptive fields, which has a greater advantage over increasing the number of blocks. However, increasing $DF$ and block number eventually brings only limited improvement at the cost of slower speed. Thus, to balance the trade-off between performance and speed, we choose depth $DF = 4$ and $L = 10$ as our default setting.

\subsubsection{Parameter reduction}
We next design an experiment that defines \textit{all} deep learning based methods to have a similar number of parameters. Note that this keeps the respective network structures unchanged. Table \ref{tab.Ablation} shows the respective parameter numbers and a quantitative comparison on the \emph{Rain100H} dataset. As is clear, the improvement of our model becomes more significant when all methods have similar number of parameter. Our combination of dilated convolutions with a tree-structured fusion process can more efficiently represent and remove rain from images with a relatively lightweight network architecture.
\begin{table}[t]
\centering
\caption{Quantitative comparisons on parameters reduction.}
\resizebox{\linewidth}{!}{
\label{tab.Ablation}
\begin{tabular}{|c|c|c|c|c|c|c|}
\hline
              &DDN    &JORDER &DID    &RESCAN &Ours   \\
\hline
SSIM          &0.79   &0.81   &0.82   &0.84   &0.88   \\
\hline
PSNR          &25.14  &25.47  &25.65  &26.17  &27.46  \\
\hline
Parameters \# &33,267 &36,528 &35,812 &34,790 &35,427 \\
\hline
\end{tabular}
}
\end{table}

\subsubsection{Loss function}
\begin{figure}
\begin{center}
\subfigure[Input]{\includegraphics[width = 0.77in,height = 0.77in]{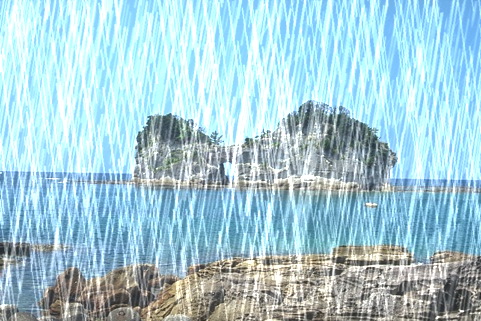}}
\subfigure[MSE loss]{\includegraphics[width = 0.77in,height = 0.77in]{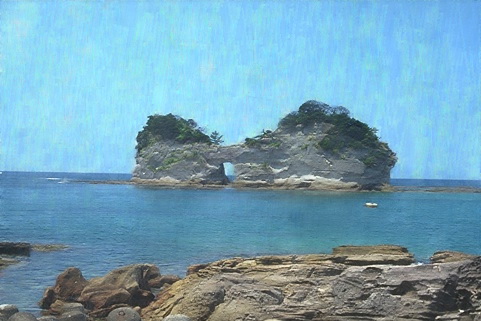}}
\subfigure[SSIM loss]{\includegraphics[width = 0.77in,height = 0.77in]{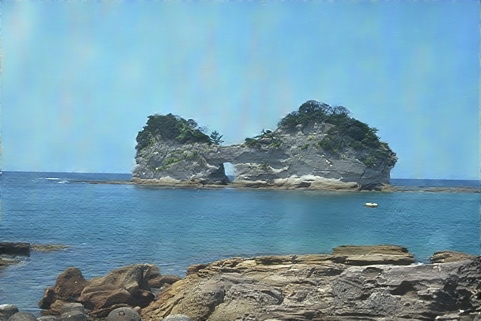}}
\subfigure[Eq. (\ref{eq.loss})]{\includegraphics[width = 0.77in,height = 0.77in]{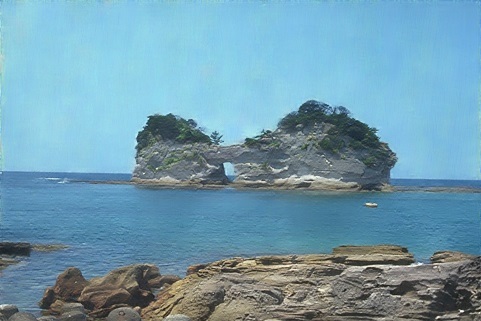}}
\caption{An example by using different losses. Using SSIM + MSE loss generates a clean result with good global contrast.}
\label{fig.loss}
\end{center}
\end{figure}

We also test the impact of using the MSE and SSIM loss functions. Figure \ref{fig.loss} shows one visual comparison on the \emph{Rain100H} dataset. As shown in Figure \ref{fig.loss}(b), using only MSE loss generates an overly-smooth image with obvious artifacts, because the $\ell_2$ penalty over-penalizes larger errors, which tend to occur at edges. SSIM focuses on structural similarity and is appropriate for preserving details. However, the result has a low contrast as shown in Figure \ref{fig.loss}(c). This is because the image contrast is related to low-frequency information, which is not penalized as heavily by SSIM. Using the combined loss in Equation (\ref{eq.loss}) can further improve the performance. In Table \ref{tab.differentloss}, we show the quantitative evaluations of using different loss functions. We observe that PSNR is a function of MSE, and so in both cases the quantitative performance measure should favor the objective function that optimizes over this measure. Figure \ref{fig.loss} shows that this balance results in a better image.

\begin{table}[t]
\caption{Quantitative comparisons for different loss functions.}
\resizebox{\linewidth}{!}{
\begin{tabular}{|*{7}{c|}}
\hline
Loss              & \multicolumn{2}{|c|}{MSE} & \multicolumn{2}{|c|}{SSIM} & \multicolumn{2}{|c|}{MSE + SSIM} \\\cline{1-7}
                  & SSIM &  PSNR              & SSIM  & PSNR                                    & SSIM & PSNR  \\\hline
\emph{Rain100H}   & 0.85 &  28.63             & 0.88  & 25.47                                   & 0.88 &27.46\\
\hline
\emph{Rain1400}   & 0.91 &  33.24             & 0.92  & 30.41                                   & 0.92 &31.32\\
\hline
\emph{Rain1200}   & 0.90 &  33.21             & 0.93  & 30.13                                   & 0.92 &32.30\\
\hline
\end{tabular}
\label{tab.differentloss}
}
\end{table}

\begin{figure}[t!]
\begin{center}
\subfigure[ResNet (shallow)]{\includegraphics[width = 1.5in]{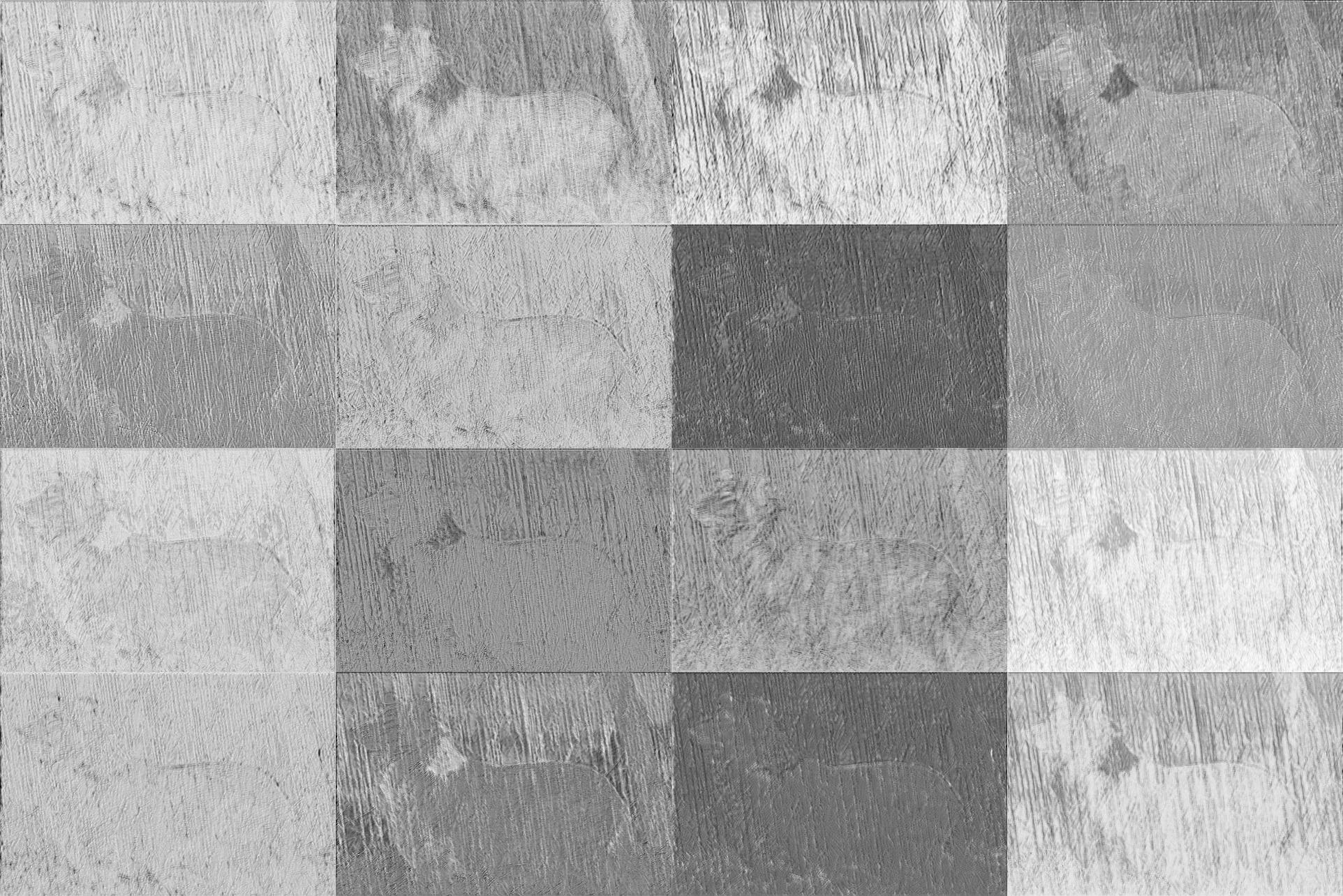}}
\subfigure[DenseNet (shallow)]{\includegraphics[width = 1.5in]{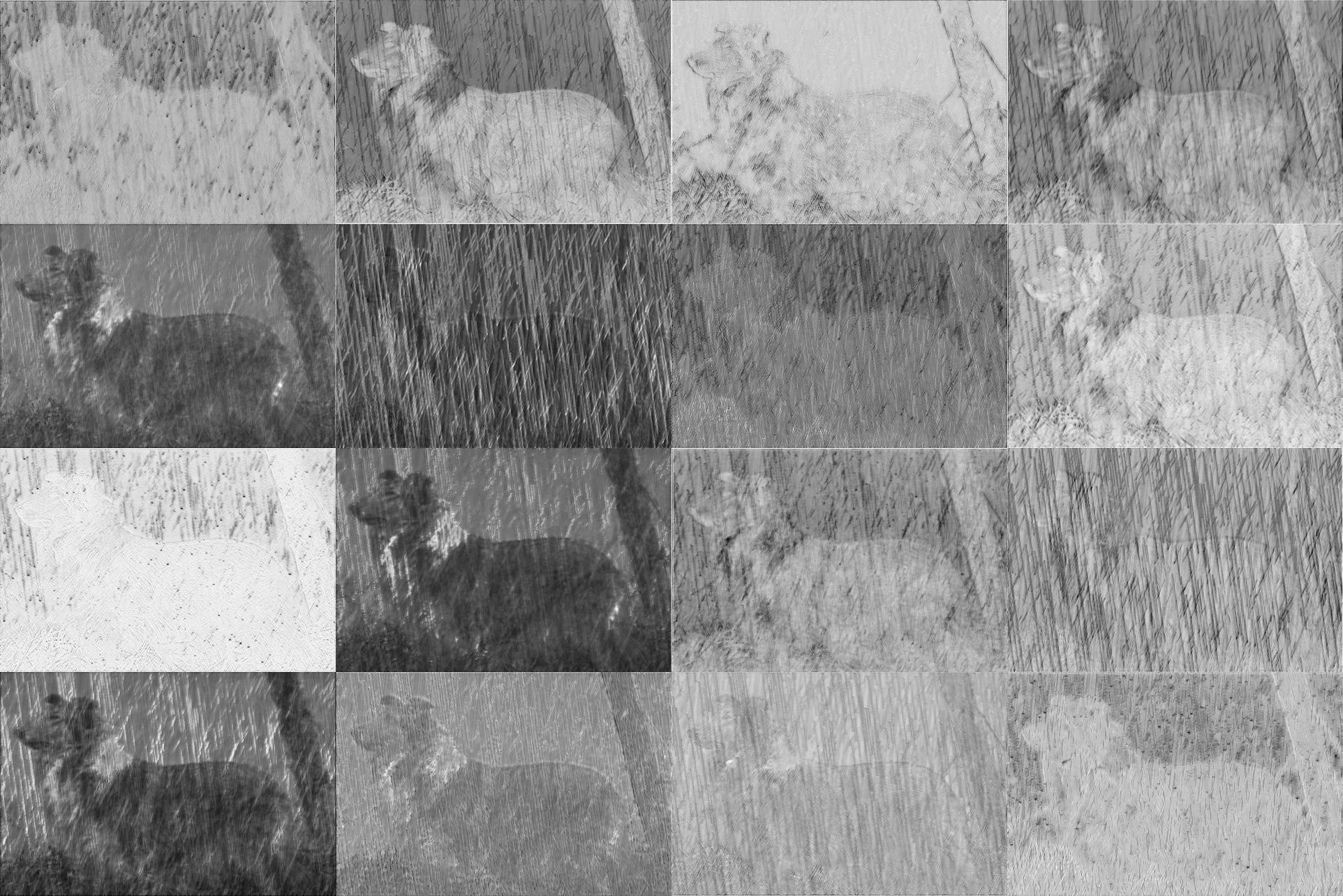}} \\
\subfigure[ResNet (deep)]{\includegraphics[width = 1.5in]{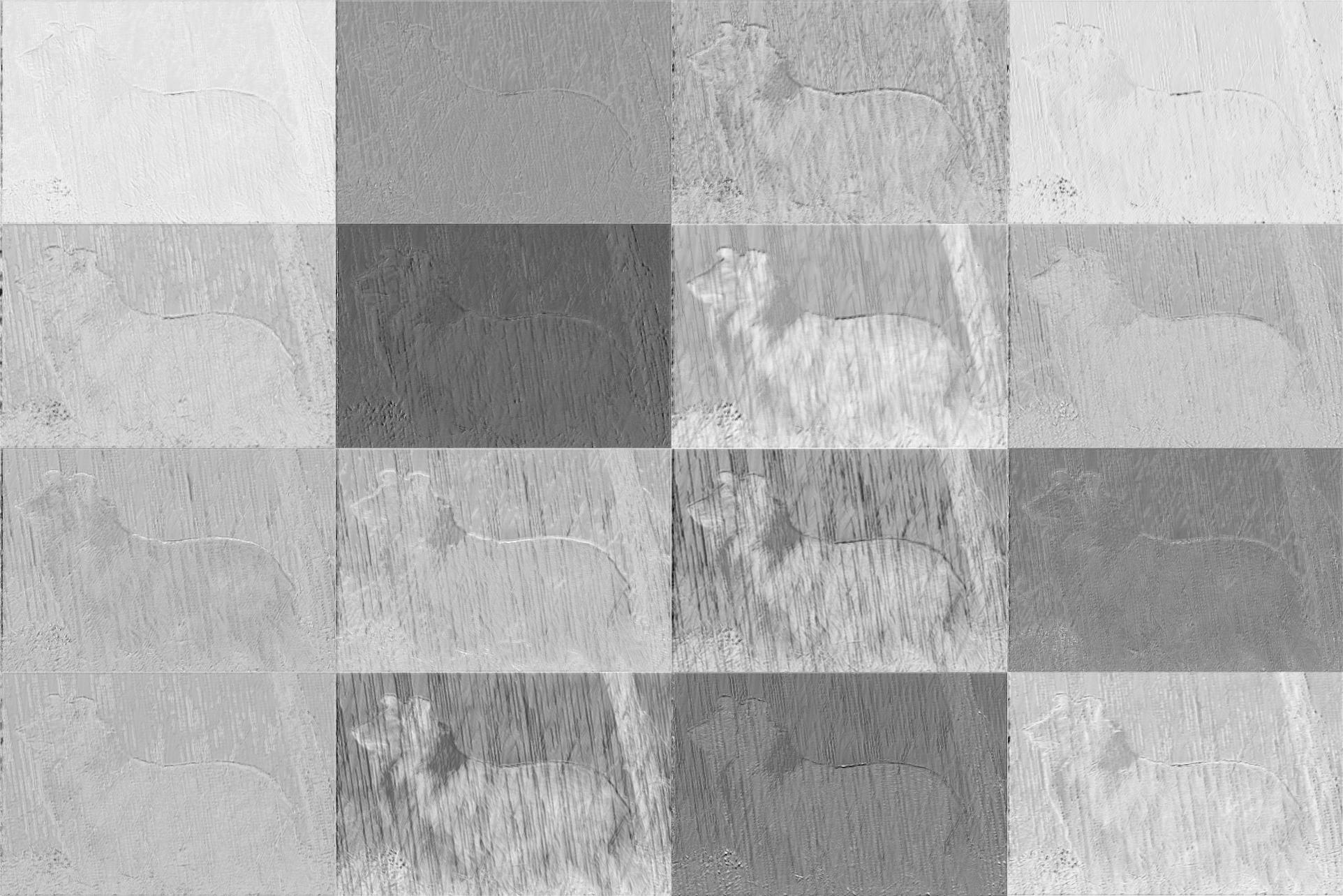}}
\subfigure[DenseNet (deep)]{\includegraphics[width = 1.5in]{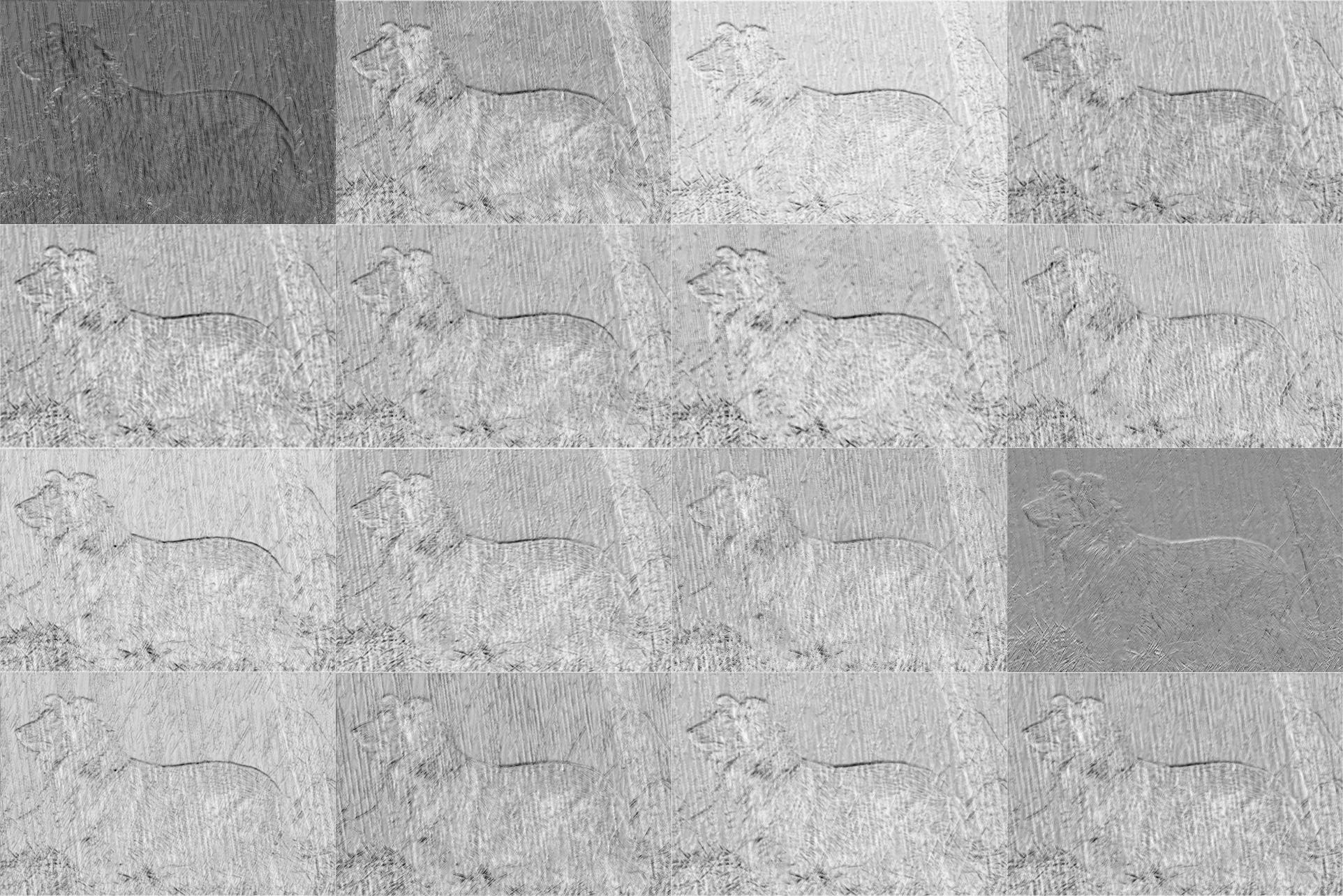}} \\
\subfigure[Our]{\includegraphics[width = 1.5in]{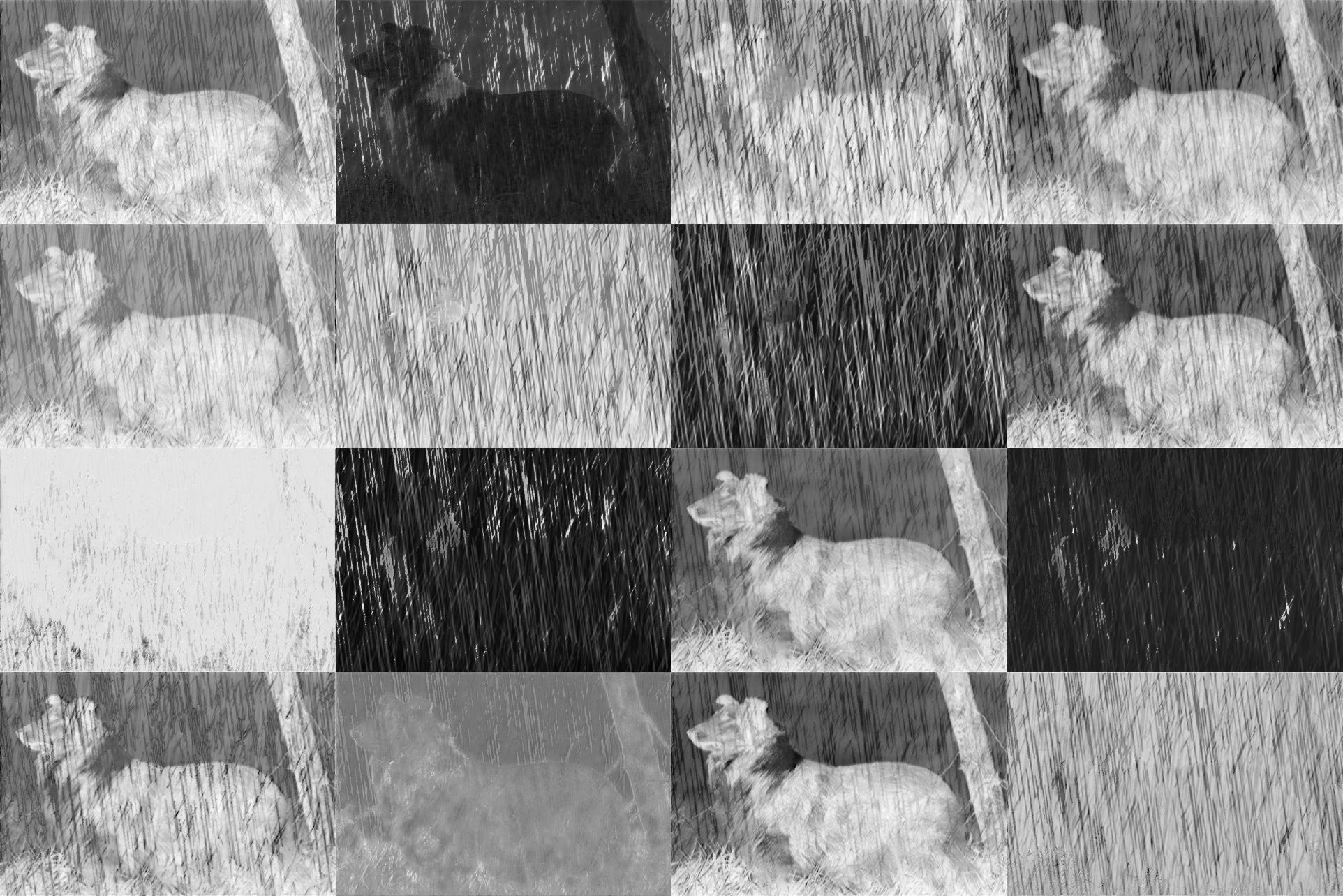}}
\caption{Visualizations of feature maps of last convolutional layers of ResNet \cite{he2016deep}, DenseNet \cite{huang2017densely} and our model.}
\label{fig.RES}
\end{center}
\end{figure}
\subsection{Comparison with ResNet and DenseNet}
Finally, to further support our tree-structured feature fusion strategy, we compare with two popular general network architectures, i.e., ResNet \cite{he2016deep} and DenseNet \cite{huang2017densely}. These methods are also designed with feature aggregation in mind. We use the same hyper-parameter setting of kernel number to 16, and use the same loss function (\ref{eq.loss}). We remove all pooling operations for this image regression problem and performance is evaluated on the \emph{Rain100H} dataset.

First, we build relatively shallow models based on ResNet (10 convolutional layers) and DenseNet (3 dense blocks), to enforce all networks to have roughly the same number of parameters. Then, we increase the depth of ResNet (82 convolutional layers) and DenseNet (20 dense blocks) to construct deeper models, which is a common means for significantly increasing model capacity and improving performance with these models \cite{tai2017memnet}. As shown in Table \ref{tab.ResNet}, our model significantly outperforms other two models under similar parameter number settings. Moreover, deeper models only achieve marginal improvement at the cost of parameter increase and computational burden.

We also show an visual comparison of the feature maps of the last convolutional layers in Figure \ref{fig.RES}. Intuitively, deeper networks can generate richer representations due to more nonlinear operations and information propagation \cite{huang2017densely,tai2017memnet}. However, as shown in Figure \ref{fig.RES}, even when compared to the deepest features, our relatively shallow tree-structured model can generate more distinguishing features. Both rain streaks and objects are represented and highlighted well. We believe this is because as the ResNet and DenseNet deepen, direct and one-way feature propagation can only reduce a limited amount of redundancy, which hinders new features generation. Our model, on the other hand, reduces feature redundancy hierarchically using a tree-structured representation, which as seen in models such as wavelet trees is a good method for representing redundant information within an image.

\begin{table}[t]
\caption{Comparisons with ResNet and DenseNet on \emph{Rain100H}.}
\resizebox{\linewidth}{!}{
\begin{tabular}{|*{6}{c|}}
\hline
               &\multicolumn{2}{|c|}{ResNet} & \multicolumn{2}{|c|}{DenseNet} & our     \\\cline{1-6}
               & shallow &  deep             & shallow  & deep                & -       \\
\hline
SSIM           & 0.84    &  0.89             & 0.85     & 0.90                & 0.88    \\
\hline
PSNR           & 25.71   &  28.03            & 25.87    & 28.24               & 27.46    \\
\hline
Parameters \#  & 38,003  &  186,483          & 35,683   & 232,883             & 35,427    \\
\hline
\end{tabular}
\label{tab.ResNet}
}
\end{table}

\section{Conclusion}
We have introduced a deep tree-structured fusion model for single image deraining. By using a simple feature fusion operation in the network, both spatial and content information are fused to reduce redundant information. These new and compact fused features leads to a significant improvement on both synthetic and real-world rainy images with a significantly reduced number of parameters We anticipate that this tree-structured framework can improve other vision and image restoration tasks and plan to investigate this in future work.

{\small
\bibliographystyle{ieee}
\bibliography{egbib}

\begin{thebibliography}{10}\itemsep=-1pt

\bibitem{abadi2016tensorflow}
M.~Abadi, A.~Agarwal, P.~Barham, et~al.
\newblock Tensorflow: Large-scale machine learning on heterogeneous distributed
  systems.
\newblock {\em arXiv preprint arXiv: 1603.04467}, 2016.

\bibitem{barnum2010analysis}
P.~C. Barnum, S.~Narasimhan, and T.~Kanade.
\newblock Analysis of rain and snow in frequency space.
\newblock {\em Int'l. J. Computer Vision}, 86(2):256--274, 2010.

\bibitem{bossu2011rain}
J.~Bossu, N.~Hautiere, and J.~P. Tarel.
\newblock Rain or snow detection in image sequences through use of a histogram
  of orientation of streaks.
\newblock {\em Int'l. J. Computer Vision}, 93(3):348--367, 2011.

\bibitem{chang2017transformed}
Y.~Chang, L.~Yan, and S.~Zhong.
\newblock Transformed low-rank model for line pattern noise removal.
\newblock In {\em ICCV}, 2017.

\bibitem{Chen2018RobustCVPR}
J.~Chen, C.~Tan, J.~Hou, L.~Chau, and H.~Li.
\newblock Robust video content alignment and compensation for rain removal in a
  {CNN} framework.
\newblock In {\em CVPR}, 2018.

\bibitem{chen2013generalized}
Y.~L. Chen and C.~T. Hsu.
\newblock A generalized low-rank appearance model for spatio-temporally
  correlated rain streaks.
\newblock In {\em ICCV}, 2013.

\bibitem{dong2016image}
C.~Dong, C.~C. Loy, K.~He, and X.~Tang.
\newblock Image super-resolution using deep convolutional networks.
\newblock {\em IEEE Trans. Pattern Anal. Mach. Intell.}, 38(2):295--307, 2016.

\bibitem{eigen2013restoring}
D.~Eigen, D.~Krishnan, and R.~Fergus.
\newblock Restoring an image taken through a window covered with dirt or rain.
\newblock In {\em ICCV}, 2013.

\bibitem{fu2017clearing}
X.~Fu, J.~Huang, X.~Ding, Y.~Liao, and J.~Paisley.
\newblock Clearing the skies: A deep network architecture for single-image rain
  removal.
\newblock {\em IEEE Trans. Image Process.}, 26(6):2944--2956, 2017.

\bibitem{fu2017removing}
X.~Fu, J.~Huang, D.~Zeng, Y.~Huang, X.~Ding, and J.~Paisley.
\newblock Removing rain from single images via a deep detail network.
\newblock In {\em CVPR}, 2017.

\bibitem{garg2007vision}
K.~Garg and S.~K. Nayar.
\newblock Vision and rain.
\newblock {\em Int'l. J. Computer Vision}, 75(1):3--27, 2007.

\bibitem{goodfellow2014generative}
I.~Goodfellow, J.~Pouget-Abadie, M.~Mirza, B.~Xu, D.~Warde-Farley, S.~Ozair,
  A.~Courville, and Y.~Bengio.
\newblock Generative adversarial nets.
\newblock In {\em NIPS}, 2014.

\bibitem{gu2017joint}
S.~Gu, D.~Meng, W.~Zuo, and L.~Zhang.
\newblock Joint convolutional analysis and synthesis sparse representation for
  single image layer separation.
\newblock In {\em ICCV}, 2017.

\bibitem{he2016deep}
K.~He, X.~Zhang, S.~Ren, and J.~Sun.
\newblock Deep residual learning for image recognition.
\newblock In {\em CVPR}, 2016.

\bibitem{Hu2018CVPR}
J.~Hu, L.~Shen, and G.~Sun.
\newblock Squeeze-and-excitation networks.
\newblock In {\em CVPR}, 2018.

\bibitem{Huang2014Self}
D.~A. Huang, L.~W. Kang, Y.~C.~F. Wang, and C.~W. Lin.
\newblock Self-learning based image decomposition with applications to single
  image denoising.
\newblock {\em {IEEE} Trans. Multimedia}, 16(1):83--93, 2014.

\bibitem{huang2017densely}
G.~Huang, Z.~Liu, L.~Van Der~Maaten, and K.~Q. Weinberger.
\newblock Densely connected convolutional networks.
\newblock In {\em CVPR}, 2017.

\bibitem{Jiang2017CVPR}
T.-X. Jiang, T.-Z. Huang, X.-L. Zhao, L.-J. Deng, and Y.~Wang.
\newblock A novel tensor-based video rain streaks removal approach via
  utilizing discriminatively intrinsic priors.
\newblock In {\em CVPR}, 2017.

\bibitem{Kang2012automatic}
L.~W. Kang, C.~W. Lin, and Y.~H. Fu.
\newblock Automatic single image-based rain streaks removal via image
  decomposition.
\newblock {\em {IEEE} Trans. Image Process.}, 21(4):1742--1755, 2012.

\bibitem{kim2013single}
J.~H. Kim, C.~Lee, J.~Y. Sim, and C.~S. Kim.
\newblock Single-image deraining using an adaptive nonlocal means filter.
\newblock In {\em IEEE ICIP}, 2013.

\bibitem{Kingma2014Adam}
D.~P. Kingma and J.~Ba.
\newblock Adam: A method for stochastic optimization.
\newblock In {\em ICLR}, 2014.

\bibitem{Kligvasser2018CVPR}
I.~Kligvasser, T.~Rott~Shaham, and T.~Michaeli.
\newblock x{U}nit: Learning a spatial activation function for efficient image
  restoration.
\newblock In {\em CVPR}, 2018.

\bibitem{krizhevsky2012imagenet}
A.~Krizhevsky, I.~Sutskever, and G.~E. Hinton.
\newblock Image{N}et classification with deep convolutional neural networks.
\newblock In {\em NIPS}, 2012.

\bibitem{Li2018VideoCVPR}
M.~Li, Q.~Xie, Q.~Zhao, W.~Wei, S.~Gu, J.~Tao, and D.~Meng.
\newblock Video rain streak removal by multiscale convolutional sparse coding.
\newblock In {\em CVPR}, 2018.

\bibitem{Li2018Recurrent}
X.~Li, J.~Wu, Z.~Lin, H.~Liu, and H.~Zha.
\newblock Recurrent squeeze-and-excitation context aggregation net for single
  image deraining.
\newblock In {\em ECCV}, 2018.

\bibitem{Li2016Rain}
Y.~Li, R.~T. Tan, X.~Guo, J.~Lu, and M.~S. Brown.
\newblock Rain streak removal using layer priors.
\newblock In {\em CVPR}, 2016.

\bibitem{Luo2015Removing}
Y.~Luo, Y.~Xu, and H.~Ji.
\newblock Removing rain from a single image via discriminative sparse coding.
\newblock In {\em ICCV}, 2015.

\bibitem{Pan2018CVPR}
J.~Pan, S.~Liu, D.~Sun, et~al.
\newblock Learning dual convolutional neural networks for low-level vision.
\newblock In {\em CVPR}, 2018.

\bibitem{Qian2018CVPR}
R.~Qian, R.~T. Tan, W.~Yang, J.~Su, and J.~Liu.
\newblock Attentive generative adversarial network for raindrop removal from a
  single image.
\newblock In {\em CVPR}, 2018.

\bibitem{ren2017video}
W.~Ren, J.~Tian, Z.~Han, A.~Chan, and Y.~Tang.
\newblock Video desnowing and deraining based on matrix decomposition.
\newblock In {\em CVPR}, 2017.

\bibitem{santhaseelan2015utilizing}
V.~Santhaseelan and V.~K. Asari.
\newblock Utilizing local phase information to remove rain from video.
\newblock {\em Int'l. J. Computer Vision}, 112(1):71--89, 2015.

\bibitem{tai2017memnet}
Y.~Tai, J.~Yang, X.~Liu, and C.~Xu.
\newblock Mem{N}et: A persistent memory network for image restoration.
\newblock In {\em ICCV}, 2017.

\bibitem{Wang2017Hierarchical}
Y.~Wang, S.~Liu, C.~Chen, and B.~Zeng.
\newblock A hierarchical approach for rain or snow removing in a single color
  image.
\newblock {\em IEEE Trans. Image Process.}, 26(8):3936--3950, 2017.

\bibitem{Wang2014SSIM}
Z.~Wang, A.~C. Bovik, H.~R. Sheikh, and E.~P. Simoncelli.
\newblock Image quality assessment: From error visibility to structural
  similarity.
\newblock {\em IEEE Trans. Image Process.}, 13(4):600--612, 2004.

\bibitem{wei2017should}
W.~Wei, L.~Yi, Q.~Xie, Q.~Zhao, D.~Meng, and Z.~Xu.
\newblock Should we encode rain streaks in video as deterministic or
  stochastic?
\newblock In {\em ICCV}, 2017.

\bibitem{Yang2017Deep}
W.~Yang, R.~T. Tan, J.~Feng, J.~Liu, Z.~Guo, and S.~Yan.
\newblock Deep joint rain detection and removal from a single image.
\newblock In {\em CVPR}, 2017.

\bibitem{yu2015multi}
F.~Yu and V.~Koltun.
\newblock Multi-scale context aggregation by dilated convolutions.
\newblock In {\em ICLR}, 2015.

\bibitem{zhang2018density}
H.~Zhang and V.~Patel.
\newblock Density-aware single image de-raining using a multi-stream dense
  network.
\newblock In {\em CVPR}, 2018.

\bibitem{zhang2017image}
H.~Zhang, V.~Sindagi, and V.~M. Patel.
\newblock Image de-raining using a conditional generative adversarial network.
\newblock {\em arXiv preprint arXiv:1701.05957}, 2017.

\bibitem{zhu2017joint}
L.~Zhu, C.~W. Fu, D.~Lischinski, and P.~A. Heng.
\newblock Joint bi-layer optimization for single-image rain streak removal.
\newblock In {\em ICCV}, 2017.

\end{thebibliography}
}

\end{document}